\pdfoutput=1
\documentclass[11pt]{article}
\usepackage[final]{acl}
\usepackage{times}
\usepackage{latexsym}
\usepackage[T1]{fontenc}
\usepackage[utf8]{inputenc}
\usepackage{microtype}
\usepackage{inconsolata}
\usepackage{setspace}
\usepackage{tabularx}
\usepackage{multirow}
\usepackage{array}
\usepackage{float}
\usepackage{graphicx}
\usepackage{amsmath}
\usepackage{amssymb}
\usepackage{booktabs}
\usepackage{subcaption}
\usepackage{makecell}
\usepackage{tabularx}

\title{CWM: Controllable White-Box Meta-Prompting for Adaptive Retrieval-Augmented Generation and Reasoning Ability}

\author{
 \textbf{Keuntae Kim\textsuperscript{1}\thanks{Equal contribution}},
 \textbf{Eunhye Jeong\textsuperscript{2}\footnotemark[1]},
 \textbf{Yong Suk Choi\textsuperscript{1}\thanks{Corresponding author}}
\\
\\
 \textsuperscript{1}Department of Computer Science, Hanyang University, Seoul, Korea
 \\
 \textsuperscript{2}Department of Artificial Intelligence, Hanyang University, Seoul, Korea
\\
    \{ktkpv94, jeh0826, cys\}{@hanyang.ac.kr}
}

\begin{document}
\maketitle

\begin{abstract} 
Recently, Large Language Models (LLMs) have gained significant attention due to their strong language understanding and generation capabilities, demonstrating impressive reasoning abilities as well as effective utilization of external knowledge. Many studies have proposed methods that specialize in improving performance for individual tasks. However, ironically, only a limited number of attempts have explored general-purpose, task-agnostic methods. In this work, we present a unified framework integrating reasoning and Retrieval-Augmented Generation (RAG) tasks. We further propose Controllable White-Box Meta-Prompting (CWM), a low-cost white-box method for adaptive RAG tasks previously dominated by black-box approaches, without requiring external decision modules or multi-sampling. CWM achieves state-of-the-art performance on three adaptive RAG benchmarks across recent LLMs, including GPT-oss-20b, Qwen3-14b, and Llama3.1-8b, while also demonstrating strong generality by extending to reasoning tasks. In addition, CWM provides controllability by enabling retrieval decisions to be regulated through the manipulation of internal model signals. Our code is available at \url{https://github.com/JeongEunhye00/CWM}.
\end{abstract}
\section{Introduction}
\label{sec:intro}

Since the emergence of Large Language Models (LLMs), research has largely focused on improving their performance and expanding their applicability along two major directions. The first direction emphasizes reasoning-intensive capabilities \cite{wei2022chainofthought, wang2022selfconsistency, yao2023react, guo2025deepseekr1, wen2025rlvr}, exemplified by tasks such as mathematical problem solving and code generation. The second focuses on knowledge augmentation \cite{lewis2020rag, izacard2021fid, izacard2022atlas, yao2023react, jiang2023flare, liu2023lostmiddle}, most notably through Retrieval-Augmented Generation (RAG). LLMs have demonstrated substantial performance gains in both domains, and as models optimized for each domain have been released, they have achieved accuracy levels sufficient for deployment in real-world applications. For instance, several models released in 2025—including GPT-5.2 \cite{openai2025gpt5systemcard}, Gemini 3 \cite{google2025gemini3pro_modelcard}, GLM-4.7 \cite{zheng2024chatglm_glm4_family}, Grok 4.1 \cite{xai2025grok4_1_modelcard}, and Qwen3 \cite{xu2025qwen3_omni}—surpass 90\% accuracy on the AIME benchmark, a task that remains highly challenging even for human experts.

Despite these advances in task-specific model performance, determining how to optimally process input prompts remains a largely unresolved problem. For example, GPT-5 introduced a routing mechanism that automatically selects an appropriate model based on the input prompt. However, this approach led to degraded generation quality, prompting a subsequent policy change that reintroduced user-controlled model selection \cite{openai2025gpt5systemcard, GPT_problem1}. Such cases highlight the continued need for research on determining optimal processing strategies conditioned on input characteristics.

A representative line of work addressing input-dependent processing is adaptive RAG, which aims to determine whether retrieval should be employed for a given input \cite{SELF-RAG, Adaptive-RAG, yan2024crag}. While recent studies have actively explored this direction, existing approaches suffer from three major limitations. First, most adaptive RAG methods rely on pre-trained classifiers or external databases, making them highly dependent on specific tasks or datasets. Second, they often require multiple inference calls to exploit model responses, resulting in increased computational cost. Third, performance evaluations are typically limited to tasks requiring external knowledge, leaving the impact of adaptive RAG on non-knowledge-intensive tasks underexplored. Consequently, the overall effect of adaptive RAG on general prompt processing remains insufficiently analyzed. Addressing these limitations constitutes the primary motivation of our work.

We attribute these issues primarily to an over-reliance on black-box approaches. As black-box methods do not exploit a model’s internal states, they often rely on external classifiers or repeated inference to guide decision-making, resulting in substantial computational cost. Meanwhile, recent studies have shown that LLMs can exhibit overconfidence or bias in their own generations \cite{overconfidence3, madaan2023selfrefine, geng-etal-2024-survey, overconfidence1,  overconfidence2}. Motivated by these observations, we propose existing black-box adaptive RAG approaches with a white-box perspective that leverages internal model signals to improve decision-making.

To this end, we build upon Self-Discover, an existing meta-prompting framework in which LLMs compose reasoning structures by selecting from atomic reasoning modules, and extend it with a white-box control mechanism based on the model’s internal signals. Specifically, we introduce an atomic module for deciding whether external knowledge is required and use the model’s Self-Perplexity on the input query as a pre-generation decision signal to guide its selection. This enables retrieval decisions to be made within the meta-prompting process itself, without relying on external modules or repeated inference. Our approach (1) does not require external decision modules, (2) relies on single-sample inference, and (3) applies to both RAG and reasoning tasks.

We propose Controllable White-Box Meta-Prompting (CWM), which achieves state-of-the-art performance on adaptive RAG benchmarks \cite{hotpotqa, strategyqa, musique} and consistently outperforms Self-Discover \cite{self-discover}—a previously state-of-the-art method for reasoning tasks—on reasoning benchmarks \cite{math500, bbh, t4d} as well. Moreover, by leveraging a white-box approach, CWM enables fine-grained control over the degree of retrieval via hyperparameter setting, allowing adaptation to different models and tasks. Overall, CWM is a controllable and dynamic prompting method that not only achieves state-of-the-art performance on existing benchmarks but also generalizes across both adaptive RAG and reasoning tasks. Our main contributions are as follows:

\begin{itemize}
\item We propose a low-cost meta-prompting approach for adaptive RAG that requires neither external decision modules nor multi-sampling inference.
\item We introduce a controllable meta-prompting framework that integrates white-box signals, enabling task- and model-specific control over prompt processing.
\item We present a unified evaluation framework that jointly assesses adaptive RAG and reasoning tasks, enabling a more general analysis of prompt processing strategies.
\end{itemize}

\begin{figure}[t]
  \includegraphics[width=\columnwidth]{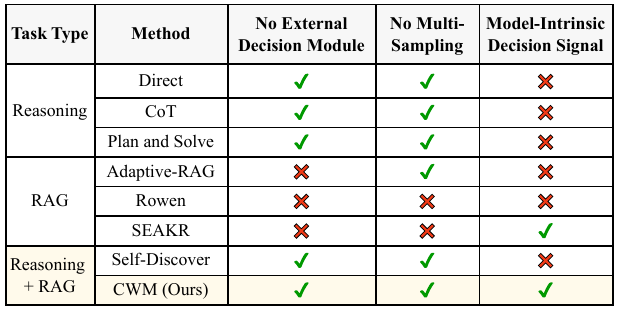}
  \caption {Overview of adaptive RAG methods and reasoning-task approaches. \textbf{No External Decision Module} indicates that the method does not rely on classifiers, databases, or external models. \textbf{No Multi-Sampling} denotes that the method does not generate multiple outputs for a single input instance. \textbf{Model-Intrinsic Decision Signal} refers to a white-box approach that leverages internal signals of the model.}
\end{figure}

\begin{figure*}[t]
  \includegraphics[width=\linewidth]{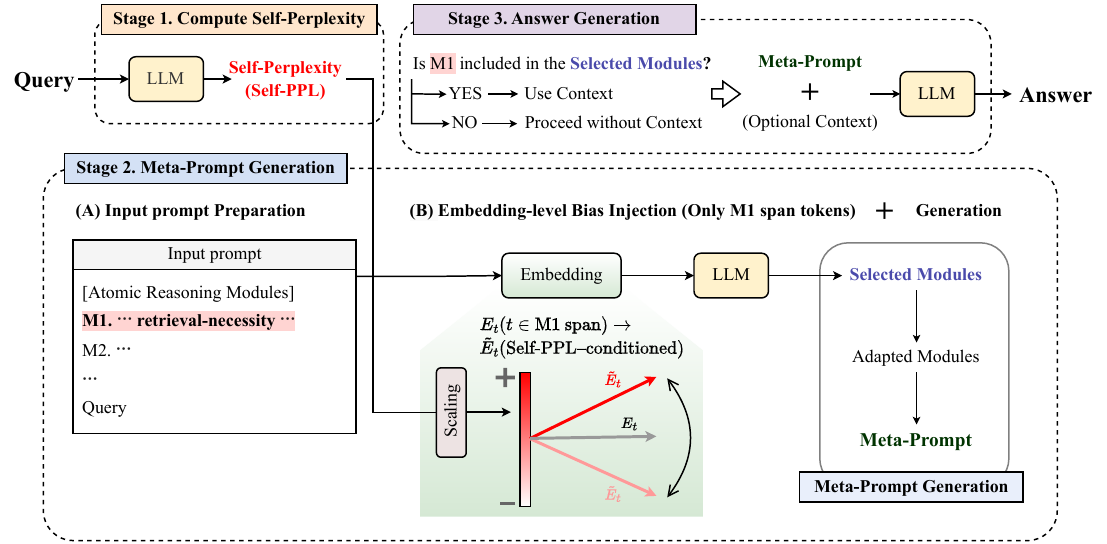}
  \caption{Overview of CWM. \textbf{Stage 1} computes the model’s self-perplexity (self-ppl) for the input query. \textbf{Stage 2} reflects self-ppl in the \textbf{M1} atomic reasoning module to decide whether external retrieval is needed, constructing a meta-prompt. \textbf{Stage 3} generates the final response conditioned on the meta-prompt and the context knowledge.}
  \label{fig1:framework}
\end{figure*}

\section{Related Work}
\paragraph{Adaptive RAG}
Adaptive Retrieval-Augmented Generation (Adaptive RAG) is a task that determines whether document retrieval should be performed based on the complexity of a given query, and thus plays a crucial role in deciding whether a model should rely on external knowledge.
Prior work has explored this problem from several perspectives. Previous study \cite{mallen2023not} proposed a method that decides whether to apply RAG based on query complexity. Adaptive-RAG \cite{Adaptive-RAG} introduced an approach that pre-trains a classifier to categorize queries into straightforward, simple, and complex types, and applies retrieval selectively based on this classification. Rowen \cite{ROWEN} proposed a consistency-based detection module that constructs cross-lingual, perturbed, and cross-model questions, and determines the necessity of retrieval by analyzing multiple generated responses. Finally, SEAKR \cite{SEAKR} generated a brief rationale for each query and used the uncertainty of the generated rationale to decide whether external knowledge retrieval is required.

\paragraph{Prompting for Reasoning Task}
Prompting strategies for reasoning tasks have been extensively studied, largely motivated by the remarkable success of Chain-of-Thought prompting \cite{wei2022chainofthought}. In particular, a growing body of work has focused on designing meta-prompts that aim to maximize the reasoning capabilities of LLMs \cite{meta-prompting1, meta-prompting4, liu2023pretrainpromptsurvey,  meta-prompting3}. Among these approaches, Plan-and-Solve Prompting \cite{plan-and-solve} demonstrated that explicitly constructing a plan for step-by-step processing within the prompt leads to improved reasoning performance.
Subsequently, Self-Discover \cite{self-discover} proposed a black-box meta-prompting approach in which LLMs autonomously compose task-specific meta-prompts by selecting from a set of atomic reasoning modules. By dynamically constructing different reasoning structures for different tasks, this method achieved substantial performance improvements. However, these meta-prompting techniques have been primarily limited to reasoning tasks. Whether LLM-based meta-prompting is effective for adaptive RAG tasks has not yet been systematically explored.
\section{Method}
% \subsection{Controllable White-Box Meta-Prompting (CWM)}
We extend the atomic reasoning prompt modules of Self-Discover by introducing a Retrieval decision prompt, enabling its application to adaptive RAG. To this end, we compute the Self-Perplexity (Self-PPL) of the input query and incorporate it into the Retrieval prompt during meta-prompt generation. Self-PPL serves as a pre-generation control signal for retrieval decisions and is injected into the embedding of the Retrieval prompt, where higher values encourage retrieval and lower values suppress it. Unlike prior approaches that estimate uncertainty from hidden states or logits during generation, which often require multi-sampling or iterative refinement, our method leverages input-level uncertainty to determine retrieval necessity before generation. This enables efficient retrieval routing with a single forward pass and significantly reduces inference cost.

\subsection{Stage 1: Compute Self-Perplexity of the Input Query}

\paragraph{Self-Perplexity Estimation}

Given an input query represented as a token sequence
\[
x = (x_1, x_2, \dots, x_T),
\]
we estimate its difficulty using sequence-level self-perplexity computed by a pretrained language model $P_\theta$.
We first compute the negative log-likelihood:
\[
\mathrm{NLL}(x)
= -\frac{1}{T-1} \sum_{t=1}^{T-1} \log P_\theta(x_{t+1} \mid x_{\leq t}),
\]
and define Self-Perplexity as
\[
\mathrm{PPL}(x) = \exp\big(\mathrm{NLL}(x)\big).
\]
Lower perplexity indicates that the model is confident about the input, while higher perplexity reflects increased uncertainty and problem difficulty.

\paragraph{Perplexity-Guided Control Signal}

To adaptively control the influence of retrieval-related reasoning modules, we map the Self-Perplexity value to a signed control variable $\alpha$.
We define three perplexity thresholds:
\[
\tau_{\mathrm{neg}} < \tau_{\mathrm{off}} < \tau_{\mathrm{on}},
\]
which correspond to easy, neutral, and difficult problem regimes, respectively.
% We compute an intermediate variable $t \in [-1,1]$ using a signed smoothstep schedule:
% \[
% t(\mathrm{PPL}) =
% \]
% \[
% \begin{cases}
% -1,
% & \mathrm{PPL} \leq \tau_{\mathrm{neg}}, \\[6pt]

% -1 + S\!\left(
% \frac{\ln(\mathrm{PPL}) - \ln(\tau_{\mathrm{neg}})}
% {\ln(\tau_{\mathrm{off}}) - \ln(\tau_{\mathrm{neg}})}
% \right),
% & \tau_{\mathrm{neg}} < \mathrm{PPL} < \tau_{\mathrm{off}}, \\[10pt]

% S\!\left(
% \frac{\ln(\mathrm{PPL}) - \ln(\tau_{\mathrm{off}})}
% {\ln(\tau_{\mathrm{on}}) - \ln(\tau_{\mathrm{off}})}
% \right),
% & \tau_{\mathrm{off}} \leq \mathrm{PPL} < \tau_{\mathrm{on}}, \\[10pt]

% 1,
% & \mathrm{PPL} \geq \tau_{\mathrm{on}}.
% \end{cases}
% \]

To compute an intermediate variable $t \in [-1,1]$, we use a smoothstep schedule. For $\mathrm{PPL}$ values satisfying $\tau_{\mathrm{neg}} < \mathrm{PPL} < \tau_{\mathrm{on}}$, we define a normalized value $z$:

{
\small
\[
z(\mathrm{PPL})
=
\begin{cases}
\dfrac{\ln(\mathrm{PPL})-\ln(\tau_{\mathrm{neg}})}
{\ln(\tau_{\mathrm{off}})-\ln(\tau_{\mathrm{neg}})},
& \tau_{\mathrm{neg}} < \mathrm{PPL} < \tau_{\mathrm{off}}, \\[10pt]

\dfrac{\ln(\mathrm{PPL})-\ln(\tau_{\mathrm{off}})}
{\ln(\tau_{\mathrm{on}})-\ln(\tau_{\mathrm{off}})},
& \tau_{\mathrm{off}} \le \mathrm{PPL} < \tau_{\mathrm{on}}.
\end{cases}
\]}

Using the corresponding normalized value within each interval, we define $t(\mathrm{PPL})$ as follows, with $t$ clamped to $-1$ for $\mathrm{PPL} \le \tau_{\mathrm{neg}}$ and to $1$ for $\mathrm{PPL} \ge \tau_{\mathrm{on}}$:

{
\small
    \[
    t(\mathrm{PPL})
    =
    \begin{cases}
    -1,
    & \mathrm{PPL} \le \tau_{\mathrm{neg}},  \\[6pt]
    
    -1 + S\!\left(z(\mathrm{PPL})\right),
    & \tau_{\mathrm{neg}} < \mathrm{PPL} < \tau_{\mathrm{off}},  \\[6pt]
    
    S\!\left(z(\mathrm{PPL}\right)),
    & \tau_{\mathrm{off}} \le \mathrm{PPL} < \tau_{\mathrm{on}},  \\[6pt]
    
    1,
    & \mathrm{PPL} \ge \tau_{\mathrm{on}}.
    \end{cases}
    \]
}

The final control coefficient $\alpha$ is computed with asymmetric caps:
\[
\alpha =
\begin{cases}
\alpha^{+}_{\mathrm{cap}} \cdot t, & t \geq 0, \\[4pt]
\alpha^{-}_{\mathrm{cap}} \cdot t, & t < 0,
\end{cases}
\]
where $\alpha^{+}_{\mathrm{cap}}$ and $\alpha^{-}_{\mathrm{cap}}$ denote the maximum positive and negative magnitudes, respectively.
This design allows the method to suppress retrieval prompt modules for easy problems ($\alpha < 0$) and emphasize them for difficult problems ($\alpha > 0$).

\subsection{Stage 2: Meta-Prompt Generation}

Unlike prompt-tuning approaches that introduce learnable prompt vectors into the input representation \cite{liu2025all}, CWM performs a training-free, in-place transformation of the embeddings corresponding to an existing retrieval-related span in the input prompt, as shown in Figure~\ref{fig1:framework}. This preserves the original sequence length while enabling the Self-PPL-derived signal to directly influence whether the retrieval module is selected.

\paragraph{Retrieval Keyword Span Identification}

Let the prompt contain a predefined keyword string that describes a Retrieval prompt.
During tokenization, each token is associated with a character-level offset $(s_i, e_i)$.
Given the character span $(c_{\mathrm{start}}, c_{\mathrm{end}})$ of the keyword string, we define the set of keyword tokens as
\[
\mathcal{K}
= \{\, i \mid \neg (e_i \leq c_{\mathrm{start}} \lor s_i \geq c_{\mathrm{end}}) \,\}.
\]

\paragraph{Perplexity-Adaptive Embedding Bias Injection}

Let $\mathbf{h}_i \in \mathbb{R}^d$ denote the input embedding of token $i \in \mathcal{K}$.
We modify the embeddings using two complementary mechanisms. We first apply a multiplicative scaling:
\[
\mathbf{h}_i^{(1)} = (1 + g_{\mathrm{scale}})\,\mathbf{h}_i,
\]
where
\[
g_{\mathrm{scale}}
= \mathrm{clamp}\big(
\alpha \cdot c_{\mathrm{scale}},
- g^{-}_{\mathrm{scale}},
g^{+}_{\mathrm{scale}}
\big).
\]

To prevent the effect from being canceled by layer normalization, we further inject a directional perturbation.
We compute the mean embedding of the keyword span:
\[
\mathbf{v}_{\mathrm{kw}} = \frac{1}{|\mathcal{K}|} \sum_{j \in \mathcal{K}} \mathbf{e}_j,
\]
where $\mathbf{e}_j$ denotes the token embedding of the keyword string.
We then extract the component orthogonal to $\mathbf{h}_i$:
\[
\mathbf{r}_i
= \mathbf{v}_{\mathrm{kw}}
- \frac{\mathbf{v}_{\mathrm{kw}}^\top \mathbf{h}_i}{\|\mathbf{h}_i\|^2} \mathbf{h}_i,
\quad
\mathbf{u}_i = \frac{\mathbf{r}_i}{\|\mathbf{r}_i\| + \epsilon}.
\]

The final embedding is given by
\[
\mathbf{h}'_i
= \mathbf{h}_i^{(1)} + g_{\mathrm{delta}} \cdot \mathbf{u}_i,
\]
\[
g_{\mathrm{delta}}
= \mathrm{clamp}\big(
\alpha \cdot c_{\mathrm{delta}},
- g^{-}_{\mathrm{delta}},
g^{+}_{\mathrm{delta}}
\big).
\]

The resulting coefficient $g_{\mathrm{delta}}$ is directly applied during decoding as an additive bias to the embeddings of retrieval-related keyword tokens, thereby modulating the influence of the corresponding reasoning module \ref{tab4:reasoning-modules}. For further details, please refer to the Appendix \ref{Appendix:method details}.
\begin{table*}[t]
\centering
\setlength{\tabcolsep}{3pt}
\fontsize{9}{10}\selectfont
\begin{tabular}{ccc cc cccc cc}
\toprule

\multirow{2}{*}[-2.5pt]{\textbf{Model}} & \multirow{2}{*}[-2.5pt]{\textbf{Dataset}} & \multirow{2}{*}[-2.5pt]{\makecell{\textbf{Metric}\\\textbf{(\%)}}} & 
\multicolumn{6}{c}{\textbf{Baselines}} & \multicolumn{2}{c}{\textbf{CWM (Ours)}} \\
\cmidrule(lr){4-9} 
\cmidrule(lr){10-11} 

& & 
& \textbf{w/GC} & \textbf{w/o GC} & \textbf{Adaptive-RAG} & \textbf{Rowen-CL} & \textbf{SEAKR} & \textbf{Self-Discover}
& \textbf{Best} & \textbf{AVG} \\ 

% ================= GPT-OSS-20B =================
\midrule
\multirow{8}{*}[-7pt]{\makecell{GPT\\oss\\20b}} 
& \multirow{3}{*}{HotpotQA} & RAG O
& 100.00 & 0.00 & 84.00 & 69.00 & 72.00 & 73.00 & 74.00 & 71.00 \\
& & EM 
& 72.00 & 31.00 & \underline{69.00} & 61.00 & 64.00 & 52.00 & \textbf{71.00} & 67.83 \\
& & F1 
& 84.53 & 38.26 & \textbf{82.88} & 75.71 & \underline{81.28} & 59.81 & 79.36 & 76.41 \\

\cmidrule(lr){2-11} 
& \multirow{3}{*}{MuSiQue} & RAG O
& 100.00 & 0.00 & 81.00 & 73.00 & 68.00 & 76.00 & 91.00 & 90.00 \\
& & EM 
& 58.00 & 20.00 & 51.00 & 46.00 & 43.00 & 48.00 & \textbf{56.00} & \underline{53.33} \\
& & F1 
& 67.51 & 25.55 & 56.68 & 52.69 & 50.12 & 53.24 & \textbf{63.76} & \underline{60.16} \\

\cmidrule(lr){2-11} 
& \multirow{2}{*}{StrategyQA} & RAG O
& 100.00 & 0.00 & 64.00 & 59.00 & 52.00 & 52.00 & 54.00 & 52.00 \\
& & Acc 
& 97.00 & 82.00 & 81.00 & \underline{83.00} & 77.00 & 65.00 & \textbf{85.00} & 81.75 \\

% ================= Qwen3-14B =================
\midrule
\multirow{8}{*}[-7pt]{\makecell{Qwen3\\14b}} 
& \multirow{3}{*}{HotpotQA} & RAG O
& 100.00 & 0.00 & 92.00 & 60.00 & 64.00 & 100.00 & 100.00 & 100.00 \\ 
& & EM 
& 70.00 & 32.00 & 68.00 & 57.00 & 59.00 & 77.00 & \textbf{78.00} & \underline{77.17} \\
& & F1 
& 81.95 & 40.73 & 79.55 & 68.05 & 70.15 & 83.83 & \textbf{84.96} & \underline{84.26} \\

\cmidrule(lr){2-11} 
& \multirow{3}{*}{MuSiQue} & RAG O
& 100.00 & 0.00 & 91.00 & 74.00 & 71.00 & 100.00 & 100.00 & 100.00 \\ 
& & EM 
& 59.00 & 20.00 & 44.00 & 38.00 & 40.00 & 52.00 & \textbf{52.00} & \underline{51.67} \\
& & F1 
& 65.80 & 29.72 & 56.66 & 44.82 & 45.76 & 59.64 & \textbf{60.55} & \underline{59.91} \\

\cmidrule(lr){2-11} 
& \multirow{2}{*}{StrategyQA} & RAG O
& 100.00 & 0.00 & 84.00 & 26.00 & 24.00 & 100.00 & 100.00 & 100.00 \\ 
& & Acc 
& 95.00 & 78.00 & 85.00 & 70.00 & 72.00 & 91.00 & \textbf{92.00} & \underline{91.50} \\

% ================= LLaMA-3.1-8B =================
\midrule
\multirow{8}{*}[-7pt]{\makecell{Llama\\3.1\\8b}} 
& \multirow{3}{*}{HotpotQA} & RAG O
& 100.00 & 0.00 & 94.00 & 97.00 & 99.00 & 65.00 & 84.00 & 84.00 \\
& & EM 
& 71.00 & 36.00 & 61.00 & 65.00 & \underline{67.00} & 55.00 & \textbf{68.00} & 64.17 \\
& & F1 
& 80.82 & 43.24 & 75.56 & \underline{76.28} & 78.28 & 64.45 & \textbf{76.85} & 74.13 \\

\cmidrule(lr){2-11} 
& \multirow{3}{*}{MuSiQue} & RAG O
& 100.00 & 0.00 & 68.00 & 98.00 & 94.00 & 56.00 & 91.00 & 85.00 \\
& & EM 
& 53.00 & 24.00 & 33.00 & 42.00 & 35.00 & 37.00 & \textbf{53.00} & \underline{51.42} \\
& & F1 
& 60.79 & 31.51 & 44.76 & 49.42 & 41.87 & 45.27 & \textbf{62.33} & \underline{59.25} \\

\cmidrule(lr){2-11} 
& \multirow{2}{*}{StrategyQA} & RAG O
& 100.00 & 0.00 & 44.00 & 14.00 & 22.00 & 51.00 & 73.00 & 67.00 \\
& & Acc  
& 93.00 & 75.00 & 67.00 & 76.00 & 78.00 & 70.00 & \textbf{82.00} & \underline{79.42} \\
\bottomrule
\end{tabular}
\caption{Results of adaptive RAG tasks. RAG O denotes the percentage of instances for which each method determines that retrieval is required. The \textbf{Best} and \textbf{AVG} values of CWM summarize performance across experiments with different hyperparameter settings. Performance under each hyperparameter setting is reported in Appendix Tables~\ref{tab:rag_ours_res_gpt}–\ref{tab:rag_ours_res_llama}. \textbf{Best} denotes the highest performance observed, while \textbf{AVG} denotes the mean performance across all settings. The highest and second-highest results, excluding gold-context (\textbf{GC}) baselines, are highlighted in bold and underlined, respectively.
}
% \caption{Results of adaptive RAG tasks. RAG O denotes the percentage of instances for which each method determines that retrieval is required. The best and second-best results, excluding gold-context (GC) baselines, are highlighted in bold and underlined, respectively.}
\label{tab1:rag_task_results}
\end{table*}

\section{Experiments}
\subsection{Benchmarks \& Models}
We evaluate adaptive RAG performance on HotpotQA \cite{hotpotqa}, MuSiQue \cite{musique}, and StrategyQA \cite{strategyqa}, which align with the primary objective of this work. For reasoning tasks, we benchmark against the Self-Discover \cite{self-discover} framework using selected BBH subtasks (Penguins in a Table, Formal Fallacies, and Web of Lies) \cite{bbh}, as well as T4D \cite{t4d} and MATH500 \cite{math500}. Using 100 randomly sampled instances per dataset, we conduct all experiments on GPT-oss-20b \cite{agarwal2025gptoss}, Qwen3-14b \cite{yang2025qwen3}, and Llama3.1-8b \cite{grattafiori2024llama3}, which collectively represent diverse LLM architectures and model scales suitable for white-box analysis.

\begin{figure*}[t]
    \centering
    \includegraphics[width=0.9\textwidth]{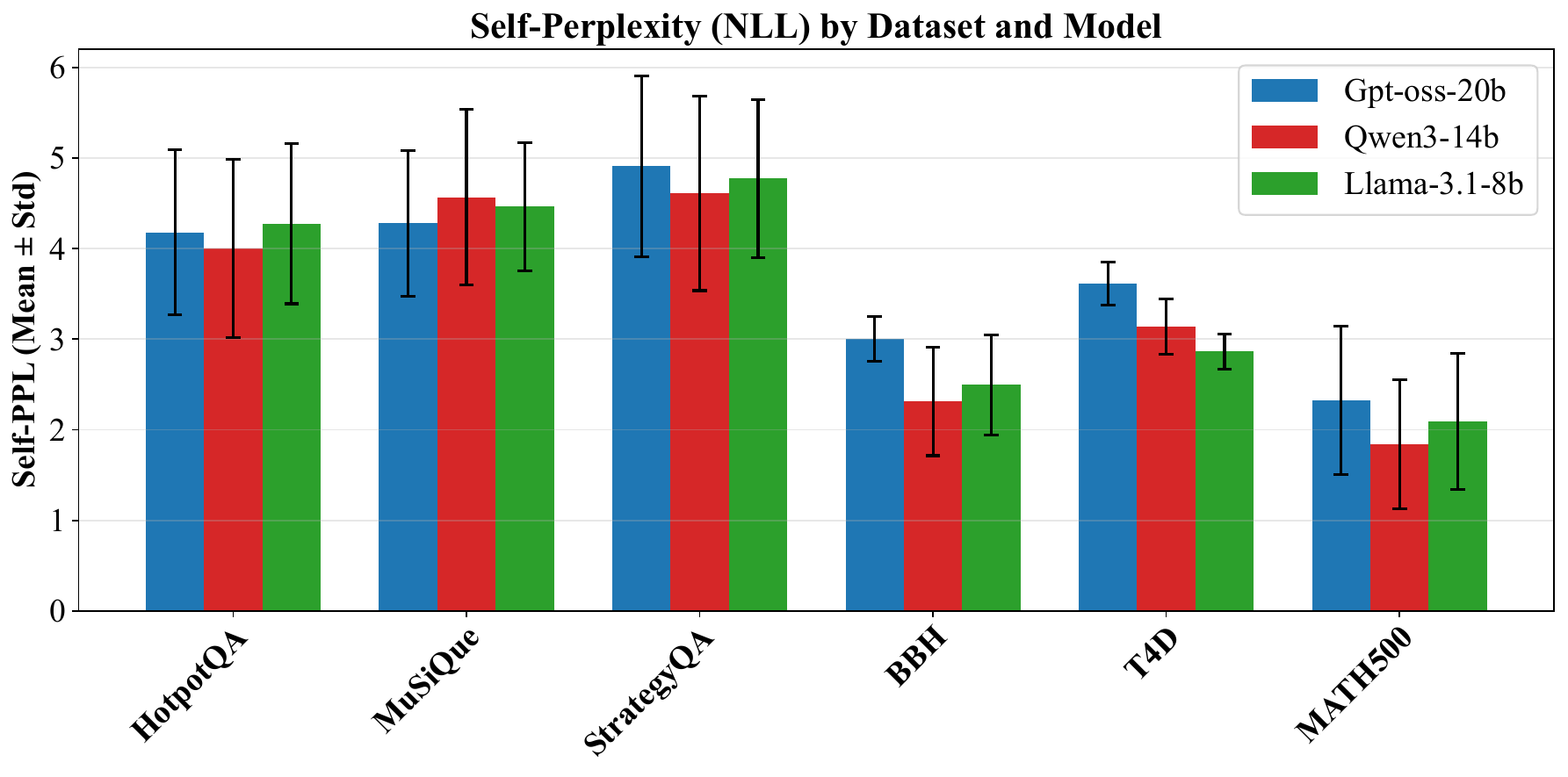}
    \caption{Evaluation of retrieval decision capability using Self-perplexity (Self-PPL). The Self-PPL values of GPT-oss-20b, Qwen3-14b, and Llama-3.1-8b were measured across three RAG benchmarks (HotpotQA, MuSiQue, StrategyQA) and three reasoning benchmarks (BBH, T4D, and MATH500).}
    \label{fig:fig3}
\end{figure*}

\begin{figure*}[t]
    \centering
    \includegraphics[width=0.9\textwidth]{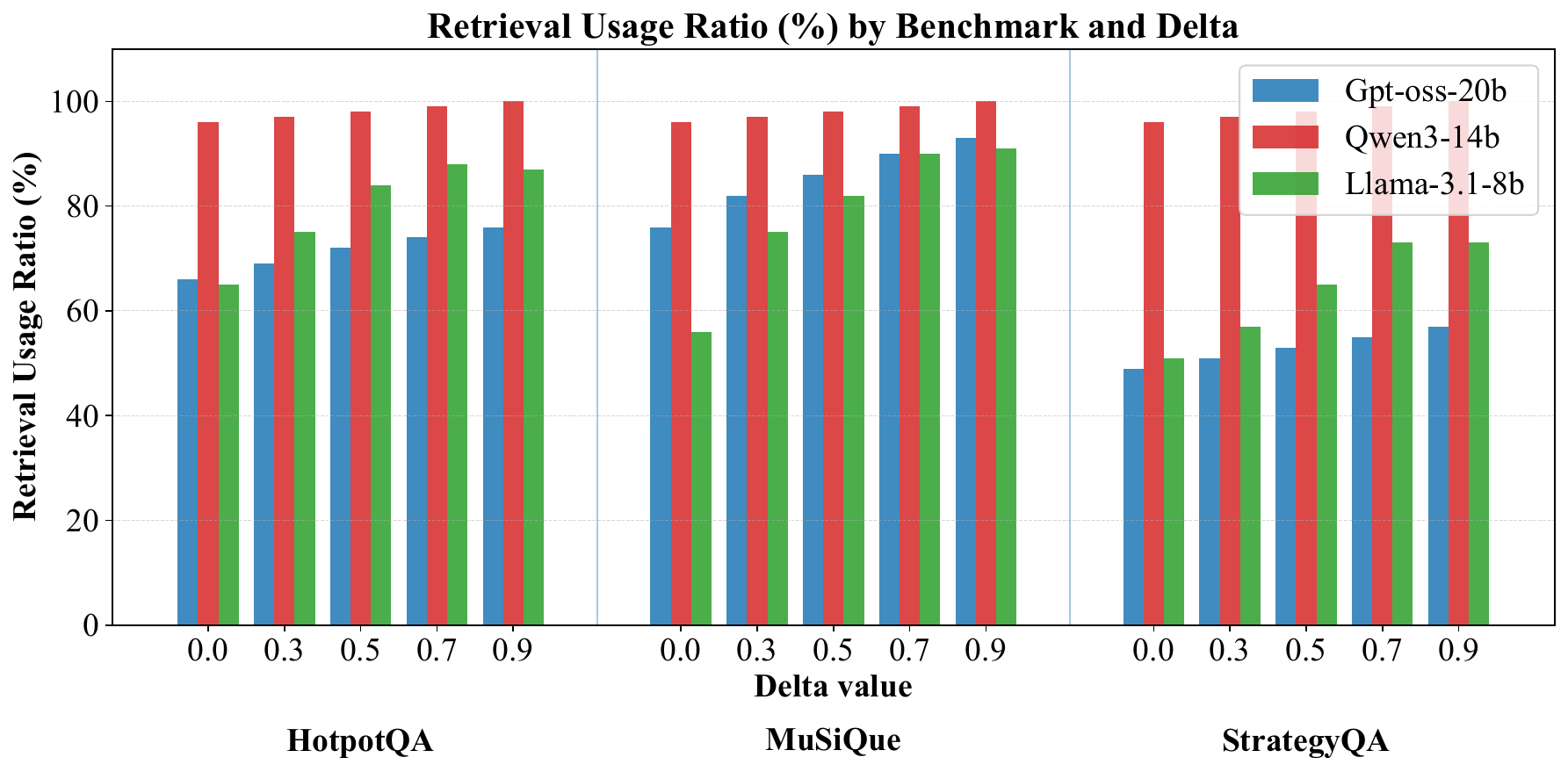}
    \caption{Comparison of retrieval usage ratios across different delta values. Experiments were conducted on GPT-oss-20b, Qwen3-14b, and Llama-3.1-8b by varying the delta parameter on HotpotQA, MuSiQue, and StrategyQA.}
    \label{fig:fig4}
\end{figure*}

\subsection{Baselines}
To assess adaptive RAG task capabilities, we benchmark against representative adaptive RAG methods, namely Adaptive-RAG \cite{Adaptive-RAG}, Rowen-CL \cite{ROWEN}, and SEAKR \cite{SEAKR}. We additionally construct a Self-Discover \cite{self-discover}-based baseline by extending its reasoning prompt modules with an explicit retrieval-decision component. To contextualize performance, we include oracle baselines with Gold Context (w/ GC) and without Gold Context (w/o GC), which provide empirical upper and lower bounds. Finally, we evaluate all methods on the three reasoning benchmarks used in Self-Discover to isolate the impact of incorporating white-box signals beyond black-box prompting.

\subsection{Implementation Details}
All outputs were generated using greedy decoding with a maximum output length of 1024 tokens (\texttt{temperature = 0}). For GPT-oss-20b, we set the \texttt{reasoning\_level} to \texttt{low}, while for Qwen3-14b, we disabled the thinking mode by setting \texttt{enable\_thinking=False}. In the adaptive RAG task setting, when retrieval was required, both the baselines and CWM were provided with the same gold context for answer generation. All experiments were conducted using four NVIDIA RTX A5000 GPUs.

\section{Results \& Analysis}
\subsection{Results for Adaptive RAG task}
We present the experimental results on adaptive RAG tasks in Table~\ref{tab1:rag_task_results}. Across all experiments, CWM consistently achieves the best performance. Compared to Adaptive-RAG, which pre-trains a classifier on a subset of the dataset, and Rowen-CL and SEAKR, which rely on multi-sampling per instance to determine whether retrieval is required, CWM achieves strong performance with significantly lower computational cost. Moreover, when compared to the primary baseline, Self-Discover, the benefits of leveraging internal model signals become evident. While Self-Discover relies solely on meta-prompting, CWM uses the same framework but augments it with internal model signals, and consistently outperforms it across all settings.

When analyzing method-specific behaviors, Rowen-CL and SEAKR, which determine retrieval necessity through partial rationale generation or multi-sampling, exhibit relatively low RAG O ratios on StrategyQA, a comparatively easier benchmark. This suggests that these methods may attain high confidence from self-generated responses. In contrast, methods that rely on a single query, including Self-Discover and CWM, as well as Adaptive-RAG, show substantially higher RAG O ratios. Notably, Rowen-CL and SEAKR occasionally perform worse than the lower bound (w/o GC) on StrategyQA, indicating that multi-sampling can induce overconfidence and fail to identify queries that truly require retrieval.

In contrast, CWM effectively leverages atomic reasoning modules and consistently achieves performance above the lower bound across all cases. Moreover, in certain model and benchmark combinations, specifically HotpotQA with Qwen3-14b and MuSiQue with Llama-3.1-8b, CWM even surpasses the gold-context baseline (w/ GC). CWM also exhibits the highest RAG O ratios across all models on MuSiQue, which is the most challenging benchmark among the three. Given that all models show very low performance under the w/o GC setting on MuSiQue, this behavior can be considered both desirable and meaningful.

\subsection{Empirical Validation of Self-PPL}
We demonstrate that Self-PPL serves as an effective signal for query-level decision-making in CWM and enables controllable retrieval behavior. Figure~\ref{fig:fig3} shows Self-PPL values across benchmarks. Following prior adaptive RAG setups, we examine whether Self-PPL distinguishes between RAG and non-RAG tasks. Across three RAG benchmarks (HotpotQA, MuSiQue, StrategyQA) and three reasoning benchmarks (BBH, T4D, MATH500), RAG benchmarks consistently exhibit higher Self-PPL. This suggests that Self-PPL effectively captures when retrieval is beneficial, making it a suitable signal for adaptive RAG.

Next, Figure~\ref{fig:fig4} examines the controllability of retrieval usage by varying the delta parameter. Except for Qwen3-14b, which consistently employs retrieval across all cases, both GPT-oss-20b and Llama3.1-8b exhibit increasing retrieval usage as the delta value increases. These results confirm that CWM allows users to regulate retrieval usage by adjusting delta, while accounting for both model and benchmark characteristics. 

We further examine how Self-PPL varies with question characteristics using annotations in HotpotQA and MuSiQue as proxies. As shown in Table~\ref{tab:self_ppl_analysis}, samples are divided into high and low Self-PPL groups based on the median within each dataset. In HotpotQA, where all questions have the same difficulty level, multi-step linking (Bridge) and Comparison questions show no consistent association with high or low Self-PPL across the three models. These results suggest that Self-PPL does not exhibit a consistent association with a particular question type. In contrast, in MuSiQue, high-hop questions consistently exhibit a higher proportion of high Self-PPL than low-hop questions across all models. Notably, a substantial proportion of low-hop questions are also classified as having high Self-PPL, suggesting that reasoning complexity alone does not fully explain Self-PPL. This remaining variation may be associated with the sufficiency of the model's internal knowledge for answering the query.

Additionally, Table~\ref{tab:no_rag_performance} reports the performance on the subset of samples for which each method decides not to perform retrieval. CWM achieves the best performance across all methods in this setting, suggesting that it more effectively identifies queries that can be handled without external knowledge. Overall, these findings provide additional support for the use of Self-PPL as a query-level control signal for retrieval decisions in CWM. Further validation of Self-PPL is provided in Appendix~\ref{app:self_ppl_analysis}.

\subsection{Results for Reasoning task}
In Table \ref{tab2:reasoning_task_results}, we compare the performance of CWM with Self-Discover, a method originally designed for reasoning tasks, to examine the extensibility of CWM to reasoning tasks. The results show that CWM consistently outperforms Self-Discover across all models and benchmarks. This indicates that injecting bias into the retrieval prompt can positively influence the selection of other atomic reasoning modules as well. In this study, we focus on the presence or absence of retrieval and therefore inject bias only into the retrieval prompt.

\begin{table}[t]
\centering
\small
\setlength{\tabcolsep}{3.5pt}
\begin{tabular}{clcccc}
\toprule
\multirow{2}{*}[-3pt]{\textbf{Model}} &
\multirow{2}{*}[-3pt]{\makecell{\textbf{Self-}\\\textbf{PPL}}} &
\multicolumn{2}{c}{\textbf{HotpotQA}} &
\multicolumn{2}{c}{\textbf{MuSiQue}} \\
\cmidrule(lr){3-4}
\cmidrule(lr){5-6}
& & \textbf{Bridge} & \textbf{Comparison} &
\textbf{H-hop} & \textbf{L-hop} \\

\midrule
\multirow{2}{*}{\makecell{GPT-oss\\20b}}
& High & 0.48 & 0.58 & 0.62 & 0.43 \\
& Low  & 0.52 & 0.42 & 0.38 & 0.57 \\

\midrule
\multirow{2}{*}{\makecell{Qwen3\\14b}}
& High & 0.52 & 0.42 & 0.67 & 0.38 \\
& Low  & 0.48 & 0.58 & 0.33 & 0.62 \\

\midrule
\multirow{2}{*}{\makecell{Llama3.1\\8b}}
& High & 0.51 & 0.47 & 0.60 & 0.47 \\
& Low  & 0.49 & 0.53 & 0.40 & 0.53 \\
\bottomrule
\end{tabular}

\caption{Distribution of high and low Self-PPL across question groups in HotpotQA and MuSiQue. Self-PPL is dichotomized based on the median within each dataset. Bridge and Comparison denote multi-step linking and entity comparison questions, respectively; H-hop and L-hop denote questions with $\geq3$ and $\leq2$ reasoning steps, respectively.}
\label{tab:self_ppl_analysis}
\end{table}

\begin{table}[t]
\centering
% \small
\setlength{\tabcolsep}{2.5pt}
\fontsize{8.5}{10}\selectfont
\renewcommand{\arraystretch}{1.05}
\begin{tabular}{clccc}
\toprule
\textbf{Model} & \textbf{Method} & \textbf{HotpotQA} & \textbf{MuSiQue} & \textbf{StrategyQA} \\

\midrule
\multirow{5}{*}{\makecell[c]{GPT\\oss\\20b}}
 & Adaptive-RAG  & 50.7 & 25.1 & 82.6 \\
 & Rowen-CL      & 52.4 & 26.8 & 83.9 \\
 & SEAKR         & 45.3 & 19.8 & 78.1 \\
 & Self-Discover & 48.9 & 22.6 & 80.4 \\
 & CWM   & \textbf{54.3} & \textbf{29.4} & \textbf{85.2} \\

\midrule
\multirow{5}{*}{\makecell[c]{Qwen3\\14b}}
 & Adaptive-RAG  & 55.1 & 27.9 & 85.2 \\
 & Rowen-CL      & 56.8 & 29.5 & 86.6 \\
 & SEAKR         & 49.1 & 21.7 & 81.2 \\
 & Self-Discover & 52.8 & 24.3 & 83.5 \\
 & CWM   & \textbf{57.6} & \textbf{31.7} & \textbf{87.4} \\

\midrule
\multirow{5}{*}{\makecell[c]{Llama\\3.1\\8b}}
 & Adaptive-RAG  & 44.5 & 21.4 & 74.2 \\
 & Rowen-CL      & 46.1 & 23.0 & 75.6 \\
 & SEAKR         & 38.5 & 16.9 & 69.8 \\
 & Self-Discover & 41.2 & 18.7 & 71.6 \\
 & CWM   & \textbf{50.0} & \textbf{26.0} & \textbf{76.8} \\

\bottomrule
\end{tabular}
\caption{Performance on the subset of samples where retrieval is not triggered within each benchmark.}
\label{tab:no_rag_performance}
\end{table}

\begin{table}[t]
\centering
\setlength{\tabcolsep}{2.5pt}
\renewcommand{\arraystretch}{0.95}
\fontsize{8.5}{10}\selectfont
\begin{tabular}{c cc cc cc}

\toprule
\multirow{2}{*}[-7pt]{\textbf{Model}} & \multirow{2}{*}[-7pt]{\textbf{Dataset}} & \multirow{2}{*}[-7pt]{\textbf{Metric}} & \multicolumn{2}{c}{\textbf{Baselines}} & \multicolumn{2}{c}{\textbf{CWM}} \\
\cmidrule(lr){4-5}
\cmidrule(lr){6-7}
&  &  & \textbf{Base} & \makecell{\textbf{Self-}\\\textbf{Discover}} & \textbf{Best} & \textbf{AVG} \\

% ===================== GPT OSS 20B =====================
\midrule
\multirow{6}{*}[-7pt]{\makecell{GPT\\oss\\20b}}
& \multirow{2}{*}{BBH} & M\_AVG & -     & 4.59 & 5.09 & 5.16 \\
&                       & ACC    & 72.33 & 65.67 & \textbf{78.67} & \underline{76.19} \\
\cmidrule{2-7}
& \multirow{2}{*}{T4D} & M\_AVG & -   & 5.09 & 5.57 & 5.65 \\
&                      & ACC    & 36.00  & \underline{52.00}   & \textbf{55.00}   & 47.83 \\
\cmidrule{2-7}
& \multirow{2}{*}{MATH500} & M\_AVG & -   & 4.57 & 4.32 & 4.41 \\
&                          & ACC    & \textbf{99.00}  & 87.00   & \underline{96.00}   & 92.25 \\

% ===================== Qwen3 14B =====================
\midrule
\multirow{6}{*}[-7pt]{\makecell{Qwen3\\14b}}
& \multirow{2}{*}{BBH} & M\_AVG & -     & 2.90 & 2.91 & 2.89 \\
&                       & ACC    & 67.33 & 91.00   & \textbf{92.00}   & \underline{91.50} \\
\cmidrule{2-7}
& \multirow{2}{*}{T4D} & M\_AVG & -   & 3.03 & 2.97 & 2.98 \\
&                      & ACC    & \textbf{51.00}  & 42.00   & \underline{43.00}   & 42.33 \\
\cmidrule{2-7}
& \multirow{2}{*}{MATH500} & M\_AVG & -   & 3.13 & 3.15 & 3.15 \\
&                          & ACC    & 91.00  & \underline{93.00}   & \textbf{94.00}   & 92.92 \\

% ===================== Llama 3.1 8B =====================
\midrule
\multirow{6}{*}[-7pt]{\makecell{Llama\\3.1\\8b}}
& \multirow{2}{*}{BBH} & M\_AVG & -     & 10.32 & 10.19 & 10.17 \\
&                       & ACC    & 50.33 & \underline{68.33} & \textbf{68.67} & 66.31 \\
\cmidrule{2-7}
& \multirow{2}{*}{T4D} & M\_AVG & -   & 5.41 & 5.29 & 5.33 \\
&                      & ACC    & \textbf{34.00}  & 25.00   & \underline{26.00}   & 24.25 \\
\cmidrule{2-7}
& \multirow{2}{*}{MATH500} & M\_AVG & -   & 7.65 & 7.00 & 7.49 \\
&                          & ACC    & 21.00  & \underline{56.00}   & \textbf{64.00}   & 58.92 \\

\bottomrule
\end{tabular}
\caption{Results of reasoning tasks. \textbf{M\_AVG} represents the average number of selected atomic prompt modules. For the direct method, GPT-oss-20b was evaluated using the reasoning:low setting, while Qwen3-14b was evaluated in non-thinking mode.}
\label{tab2:reasoning_task_results}
\end{table}
\begin{table}[t]
\centering
\small
\setlength{\tabcolsep}{4.5pt}
\renewcommand{\arraystretch}{1.1}
\begin{tabular}{clcc}
\toprule
\textbf{Model} & \textbf{Method} & \textbf{Time (s)} & \textbf{Inference Calls} \\
\midrule

\multirow{5}{*}{\makecell{GPT\\oss\\20b}}
& Adaptive-RAG  & $82$--$208$ $+\alpha$ & $2$--$5$ $+\alpha$ \\
& SEAKR         & 753.88                & $2$--$5$ \\
& Self-Discover & 120.15                & 3 \\
& Rowen-CL      & 833.58                & 39.69 \\
& CWM           & 121.79                & 3 \\
\midrule

\multirow{5}{*}{\makecell{Qwen3\\14b}}
& Adaptive-RAG  & $25$--$65$ $+\alpha$  & $2$--$5$ $+\alpha$ \\
& SEAKR         & 269.22                & $2$--$5$ \\
& Self-Discover & 37.34                 & 3 \\
& Rowen-CL      & 554.68                & 39.60 \\
& CWM           & 36.23                 & 3 \\
\midrule

\multirow{5}{*}{\makecell{Llama\\3.1\\8b}}
& Adaptive-RAG  & $45$--$115$ $+\alpha$ & $2$--$5$ $+\alpha$ \\
& SEAKR         & 453.56                & $2$--$5$ \\
& Self-Discover & 66.89                 & 3 \\
& Rowen-CL      & 312.26                & 39.97 \\
& CWM           & 65.59                 & 3 \\
\bottomrule
\end{tabular}
\caption{Average inference time and number of inference calls per data instance on HotpotQA.}
\label{tab3:time_calls_comparision}
\end{table}

\subsection{Analysis for Efficiency}

We report the inference time and the number of inference calls for each method and model on HotpotQA in Table~\ref{tab3:time_calls_comparision}. As several methods involve multiple processing steps depending on query complexity, we adopt HotpotQA as the benchmark due to its multi-hop reasoning requirements. For a fair comparison, we disable vLLM and FlashAttention for all methods. All reported times and inference call counts include one inference call and its corresponding time for generating the final answer.

We first discuss Adaptive-RAG. This method trains a classifier on a subset of the dataset using silver, binary labels, which requires a large number of additional model inference calls. We denote these additional costs as $+\alpha$, including both inference calls and classifier training time. Accordingly, the reported inference time and call counts excluding $+\alpha$ reflect only the test-time cost after labeling and training. Adaptive-RAG classifies each query as "Straightforward", "Simple", or "Complex" and processes them accordingly. As a result, when a query is classified as Straightforward, it achieves the lowest inference time and the fewest inference calls among all methods.

SEAKR dynamically determines whether retrieval is required and how many iterations to perform using self-aware uncertainty estimation. However, as shown in Table~\ref{tab3:time_calls_comparision}, this design leads to substantially higher inference latency in practice, despite using a relatively small number of inference calls. This overhead does not arise from repeated external invocations, but rather from the computationally expensive k-sample-based uncertainty extraction and Gram-based computations performed within a single inference. As a result, SEAKR incurs significant wall-clock latency even with a limited number of inference calls, reflecting a design choice that prioritizes uncertainty-aware decision-making over computational efficiency.

Rowen-CL is built on chain-of-thought-based iterative reasoning and inherently relies on repeatedly invoking the model to improve answer reliability. As shown in Table~\ref{tab3:time_calls_comparision}, this repeated-invocation strategy results in a large number of inference calls and correspondingly high inference latency. Importantly, the increased cost stems from the accumulation of many reasoning steps rather than the complexity of individual inference calls. While effective at enhancing reasoning quality, this design makes Rowen-CL less suitable for real-time or cost-sensitive deployment scenarios.

For comparison with Self-Discover, it is worth noting that Self-Discover and CWM share the same structural framework. Given an input query, the model performs one inference call to select prompts from atomic reasoning modules, one inference call to generate a meta-prompt, and one inference call to produce the final answer, resulting in a total of three inference calls (additional inference calls may be introduced to further improve the quality of the generated meta-prompt). The overall inference time may vary depending on the number of atomic reasoning modules selected for a given instance.

As both CWM and SEAKR are white-box methods, we compare them directly. CWM computes only the perplexity of the input query, incurring minimal overhead, whereas SEAKR performs K-step sampling over hidden states, resulting in substantially higher inference time. Although SEAKR can mitigate this overhead through parallelized execution with vLLM, such optimizations are not considered in our setup. Overall, CWM achieves efficient performance without increasing cost compared to black-box methods such as Self-Discover.
\section{Conclusion}
Our study aims to develop a unified approach for both reasoning and adaptive RAG tasks using recent open-source LLMs. CWM performs strongly across both settings while reducing computational cost compared to prior methods that rely on external classifiers, auxiliary databases, additional models, or multi-sampling. Moreover, we show that retrieval decisions can be controlled using internal model signals derived from the input query itself, rather than from signals obtained during generation. Beyond retrieval, the atomic reasoning prompts also enable more general prompt selection. These findings open up promising directions for future research, particularly in light of the growing availability of high-performing open-source language models.
\section*{Limitations}
One limitation of this work concerns the scope of the atomic reasoning module library. We conduct our experiments by adding a single Retrieval prompt to the existing library of 39 atomic reasoning modules. The original atomic reasoning modules are adopted directly from prior work without modification, as their design reflects accumulated insights from multiple research communities. Modifying these modules would require broader consensus across different domains, which is beyond the scope of this study. In this work, we do not investigate whether Self-Perplexity can be effectively used to modulate the other 39 atomic reasoning prompts. Exploring new atomic reasoning module libraries tailored for LLMs and controlling individual modules using internal model signals would be an interesting direction for future work.

Another limitation concerns the applicability of CWM to closed-source LLMs. As a white-box approach, CWM requires access to token-level likelihoods and embedding-level representations, limiting its direct applicability to API-only models and making equivalent method-level comparisons challenging. Nevertheless, CWM remains applicable to self-hosted open-weight LLMs where internal signals are accessible, particularly in enterprise or security-sensitive settings. Extending white-box control methods to broader model settings remains an important direction for future work.

Finally, Self-PPL may not always accurately reflect retrieval necessity. Queries involving recent information beyond the model's training data may exhibit low Self-PPL despite requiring retrieval, while ambiguous or unusual phrasing may yield high Self-PPL even when retrieval is unnecessary. Incorporating complementary signals, such as temporal recency or lexical characteristics, could further improve retrieval decisions in future work.

\section*{Acknowledgments}
This work was supported by the Institute of Information and communications Technology Planning and evaluation (IITP) grant (No.RS-2025-25422680, No. RS-2020-II201373), and the National Research Foundation of Korea (NRF) grant (No. RS-2025-00520618) funded by the Korean Government (MSIT).
\bibliography{myRef}
\clearpage
\appendix
\begin{figure*}[t]
  \includegraphics[width=\linewidth]{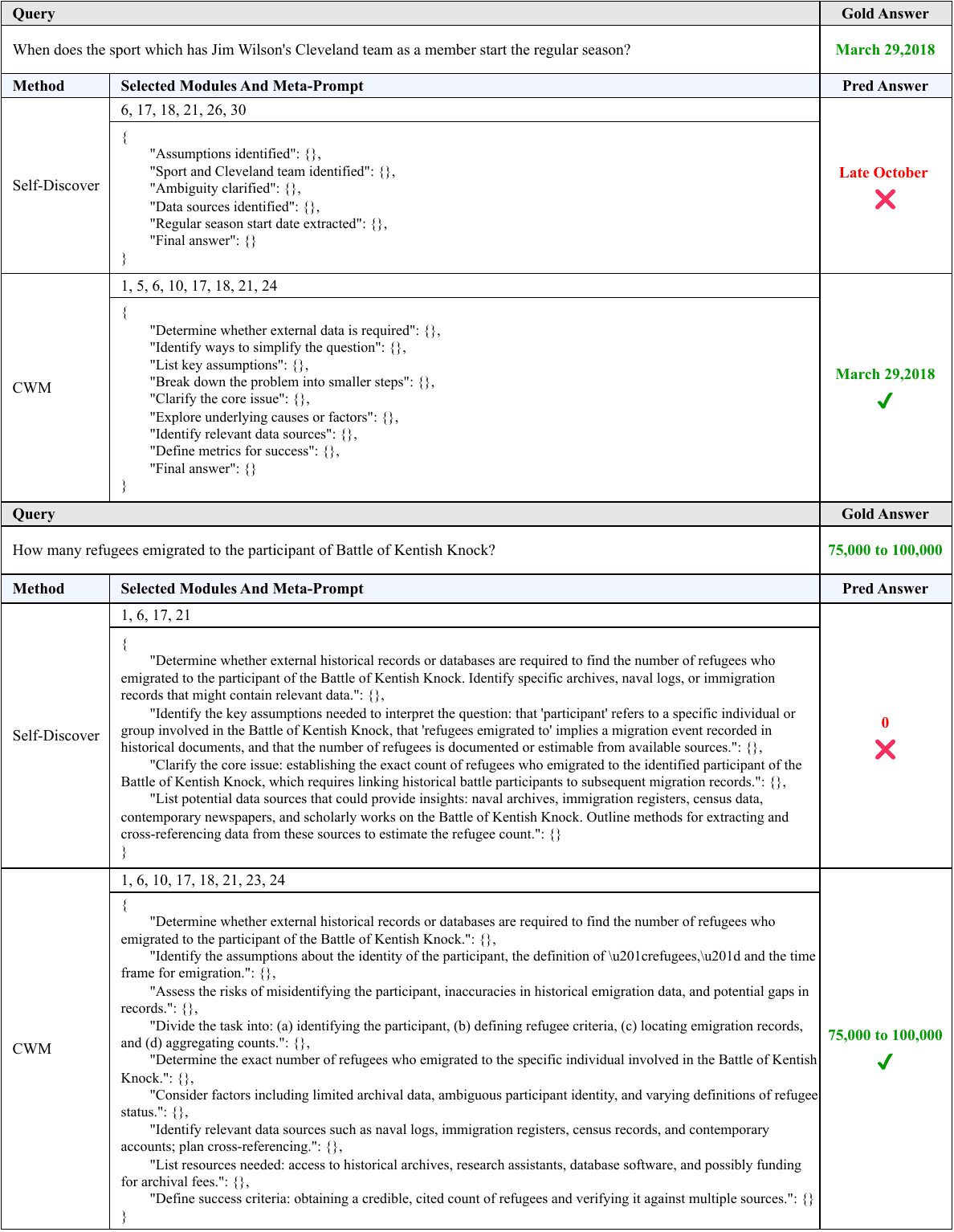}
  \caption{Comparison of selected modules and predicted answers between Self-Discover and our proposed CWM on the MuSiQue dataset using the GPT-oss-20b model.}
  \label{fig5:examples}
\end{figure*}

\begin{table*}[tp]
\centering
\setlength{\tabcolsep}{3pt}
\renewcommand{\arraystretch}{1.2}
\fontsize{9}{10}\selectfont

\begin{tabular}{@{}p{1.2em}p{0.95\linewidth}@{}}
\toprule
\multicolumn{2}{l}{\textbf{Atomic Reasoning Modules}} \\
\midrule
1  & Check if retrieval from external sources is needed to answer the question. \\
2  & How could I devise an experiment to help solve that problem? \\
3  & Make a list of ideas for solving this problem, and apply them one by one to the problem to see if any progress can be made. \\
4  & How could I measure progress on this problem? \\
5  & How can I simplify the problem so that it is easier to solve? \\
6  & What are the key assumptions underlying this problem? \\
7  & What are the potential risks and drawbacks of each solution? \\
8  & What are the alternative perspectives or viewpoints on this problem? \\
9  & What are the long-term implications of this problem and its solutions? \\
10 & How can I break down this problem into smaller, more manageable parts? \\
11 & Critical Thinking: This style involves analyzing the problem from different perspectives, questioning assumptions, and evaluating the evidence or information available. It focuses on logical reasoning, evidence-based decision-making, and identifying potential biases or flaws in thinking. \\
12 & Try creative thinking, generate innovative and out-of-the-box ideas to solve the problem. Explore unconventional solutions, thinking beyond traditional boundaries, and encouraging imagination and originality. \\
13 & Seek input and collaboration from others to solve the problem. Emphasize teamwork, open communication, and leveraging the diverse perspectives and expertise of a group to come up with effective solutions. \\
14 & Use systems thinking: Consider the problem as part of a larger system and understanding the interconnectedness of various elements. Focuses on identifying the underlying causes, feedback loops, and interdependencies that influence the problem, and developing holistic solutions that address the system as a whole. \\
15 & Use Risk Analysis: Evaluate potential risks, uncertainties, and tradeoffs associated with different solutions or approaches to a problem. Emphasize assessing the potential consequences and likelihood of success or failure, and making informed decisions based on a balanced analysis of risks and benefits. \\
16 & Use Reflective Thinking: Step back from the problem, take the time for introspection and self-reflection. Examine personal biases, assumptions, and mental models that may influence problem-solving, and being open to learning from past experiences to improve future approaches. \\
17 & What is the core issue or problem that needs to be addressed? \\
18 & What are the underlying causes or factors contributing to the problem? \\
19 & Are there any potential solutions or strategies that have been tried before? If yes, what were the outcomes and lessons learned? \\
20 & What are the potential obstacles or challenges that might arise in solving this problem? \\
21 & Are there any relevant data or information that can provide insights into the problem? If yes, what data sources are available, and how can they be analyzed? \\
22 & Are there any stakeholders or individuals who are directly affected by the problem? What are their perspectives and needs? \\
23 & What resources (financial, human, technological, etc.) are needed to tackle the problem effectively? \\
24 & How can progress or success in solving the problem be measured or evaluated? \\
25 & What indicators or metrics can be used? \\
26 & Is the problem a technical or practical one that requires a specific expertise or skill set? Or is it more of a conceptual or theoretical problem? \\
27 & Does the problem involve a physical constraint, such as limited resources, infrastructure, or space? \\
28 & Is the problem related to human behavior, such as a social, cultural, or psychological issue? \\
29 & Does the problem involve decision-making or planning, where choices need to be made under uncertainty or with competing objectives? \\
30 & Is the problem an analytical one that requires data analysis, modeling, or optimization techniques? \\
31 & Is the problem a design challenge that requires creative solutions and innovation? \\
32 & Does the problem require addressing systemic or structural issues rather than just individual instances? \\
33 & Is the problem time-sensitive or urgent, requiring immediate attention and action? \\
34 & What kinds of solution typically are produced for this kind of problem specification? \\
35 & Given the problem specification and the current best solution, have a guess about other possible solutions. \\
36 & Let’s imagine the current best solution is totally wrong, what other ways are there to think about the problem specification? \\
37 & What is the best way to modify this current best solution, given what you know about these kinds of problem specification? \\
38 & Ignoring the current best solution, create an entirely new solution to the problem. \\
39 & Let’s think step by step. \\
40 & Let’s make a step by step plan and implement it with good notion and explanation. \\
\bottomrule
\end{tabular}
\caption{Atomic reasoning modules composed of 40 high-level cognitive heuristics, including a retrieval module. Except for the retrieval module (Module 1), atomic modules 2–40 are adapted from \citet{fernando2023promptbreeder}.}
\label{tab4:reasoning-modules}
\end{table*}

\clearpage

\section[\appendixname~\thesection]{Experimental Details}
\label{Appendix_A}

\subsection[\appendixname~\thesection]{Meta-Prompting Details}
Table \ref{tab4:reasoning-modules} presents the atomic reasoning modules library used in our experiments. Following prior work~\cite{fernando2023promptbreeder, self-discover}, we construct a library of 40 atomic prompt modules by adding the prompt “Check if retrieval from external sources is needed to answer the question.” to the original set of 39 atomic reasoning modules. Since the original 39 modules were derived through consensus among multiple researchers, we do not modify them beyond the addition of the Retrieval prompt.

Figure \ref{fig5:examples} illustrates example meta-prompts generated by Self-Discover and CWM (Ours). Both Self-Discover and CWM select a subset of modules from the atomic reasoning modules library and then construct a meta-prompt based on the selected modules. The final answer is generated using the constructed meta-prompt. This process is implemented by re-constructing the procedure described in the Self-Discover \cite{self-discover}, along with corresponding coding efforts.

\subsection[\appendixname~\thesection]{Method Details}
\label{Appendix:method details}
\paragraph{Overview.}
Stage~2 of CWM injects a perplexity-derived control signal into the embeddings of
retrieval-related keyword tokens.
For the keyword-token set $\mathcal{K}$, the input embedding $\mathbf{h}_i$ of each
token $i \in \mathcal{K}$ is modified as
\begin{align}
\mathbf{h}_i^{(1)} &= (1 + g_{\mathrm{scale}})\,\mathbf{h}_i, \\
\mathbf{h}'_i
&= \mathbf{h}_i^{(1)} + g_{\mathrm{delta}}\,\mathbf{u}_i ,
\label{eq:cwm-embed-inject}
\end{align}
where $g_{\mathrm{scale}}$ and $g_{\mathrm{delta}}$ are clamped linear functions of
the control coefficient $\alpha$ (Stage~1), and $\mathbf{u}_i$ is a unit vector that
induces a directional change in $\mathbf{h}_i$.
Although Eq.~\eqref{eq:cwm-embed-inject} is model-agnostic, its effective influence
depends on the normalization layer used by the backbone LLM.

\subsubsection{Normalization Layers and Embedding Scaling}
\label{app:norm-differences}

\paragraph{LayerNorm (GPT-oss-20b).}
GPT-oss-20b employs LayerNorm, defined as
\begin{align}
\mathrm{LN}(\mathbf{x})
&= \boldsymbol{\gamma} \odot
\frac{\mathbf{x} - \mu(\mathbf{x})}{\sigma(\mathbf{x})}
+ \boldsymbol{\beta},
\label{eq:layernorm}
\end{align}
where $\mu(\mathbf{x})$ and $\sigma(\mathbf{x})$ denote the feature-wise mean and
standard deviation.
For a positive scalar $c > 0$ and $\mathbf{x}' = c \mathbf{x}$, we have
\begin{align}
\frac{\mathbf{x}' - \mu(\mathbf{x}')}{\sigma(\mathbf{x}')}
&=
\frac{c \mathbf{x} - c \mu(\mathbf{x})}{c \sigma(\mathbf{x})}
=
\frac{\mathbf{x} - \mu(\mathbf{x})}{\sigma(\mathbf{x})}.
\label{eq:ln-scale}
\end{align}
While the normalized term is invariant, the presence of residual connections and
affine parameters implies that scaling the pre-norm embedding can still affect
downstream activations in practice.

\paragraph{RMSNorm (Qwen3-14b and Llama3.1-8b).}
Qwen3-14b and Llama3.1-8b rely on RMSNorm, given by
\begin{align}
\mathrm{RMSN}(\mathbf{x})
&= \boldsymbol{\gamma} \odot
\frac{\mathbf{x}}{\mathrm{RMS}(\mathbf{x})}, \\
\mathrm{RMS}(\mathbf{x})
&= \sqrt{\frac{1}{d} \sum_{k=1}^{d} x_k^2}.
\label{eq:rmsnorm}
\end{align}
For $\mathbf{x}' = c \mathbf{x}$ with $c > 0$,
\begin{align}
\mathrm{RMSN}(\mathbf{x}')
&=
\boldsymbol{\gamma} \odot
\frac{c \mathbf{x}}{c \, \mathrm{RMS}(\mathbf{x})}
=
\mathrm{RMSN}(\mathbf{x}),
\label{eq:rmsn-scale-inv}
\end{align}
which shows that scalar embedding scaling is exactly canceled by RMSNorm.

\subsubsection{Orthogonal Delta Injection}
\label{app:orthogonal-delta}

To ensure controllability under RMSNorm, CWM introduces a directional perturbation
orthogonal to the original embedding.
Let the mean embedding of the retrieval keyword span be
\begin{align}
\mathbf{v}_{\mathrm{kw}}
&= \frac{1}{|\mathcal{K}|}
\sum_{j \in \mathcal{K}} \mathbf{e}_j ,
\label{eq:kw-mean}
\end{align}
where $\mathbf{e}_j$ denotes the token embedding of the keyword string.
The orthogonal component is computed as
\begin{align}
\mathbf{r}_i
&= \mathbf{v}_{\mathrm{kw}}
- \frac{\mathbf{v}_{\mathrm{kw}}^{\top} \mathbf{h}_i}
       {\|\mathbf{h}_i\|^2}
  \mathbf{h}_i , \\
\mathbf{u}_i
&= \frac{\mathbf{r}_i}{\|\mathbf{r}_i\| + \epsilon}.
\label{eq:orthogonal-u}
\end{align}
By construction, $\mathbf{u}_i$ is (approximately) orthogonal to $\mathbf{h}_i$.
The final perturbed embedding is
\begin{align}
\mathbf{h}'_i
&= (1 + g_{\mathrm{scale}})\,\mathbf{h}_i
+ g_{\mathrm{delta}}\,\mathbf{u}_i .
\label{eq:final-hprime}
\end{align}
Since RMSNorm normalizes magnitude but preserves direction, the directional change
introduced by $g_{\mathrm{delta}}$ survives normalization:
\begin{align}
\mathrm{RMSN}(\mathbf{h}'_i)
&= \boldsymbol{\gamma} \odot
\frac{\mathbf{h}'_i}{\mathrm{RMS}(\mathbf{h}'_i)}
\neq
\boldsymbol{\gamma} \odot
\frac{\mathbf{h}_i}{\mathrm{RMS}(\mathbf{h}_i)} .
\label{eq:rmsn-direction-change}
\end{align}

\subsubsection{GPT-oss-20b: Inference and Decoding Constraints}
\label{app:gptoss-protocol}

\paragraph{Stable attention implementation.}
CWM directly intervenes on input embeddings, which can interact unpredictably with
highly optimized attention kernels.
For GPT-oss-20b, we therefore employ a standard (eager/math) attention implementation
and disable flash-style SDP backends, prioritizing numerical stability over
throughput.

\paragraph{Reasoning verbosity control.}
GPT-oss provides a system-level control over reasoning verbosity.
We fix this setting to a low level (e.g., \texttt{Reasoning: low}) to encourage concise
outputs and to ensure that changes in the meta-prompt are primarily driven by the
CWM control signal rather than stylistic variability.

\paragraph{Protocol-aware termination and decoding.}
GPT-oss follows a structured output protocol in which the assistant’s final response
is emitted within a dedicated \emph{final} channel.
Accordingly, generation uses a protocol-specific set of stop tokens, and decoded text
is post-processed to extract only the final-channel content, ensuring consistent
evaluation.

\subsubsection{Backbone-Specific Summary}
\label{app:backbone-summary}

\paragraph{GPT-oss-20b.}
With LayerNorm, pre-norm embedding interventions are not strictly eliminated in
practice, allowing both scalar scaling ($g_{\mathrm{scale}}$) and directional
perturbation ($g_{\mathrm{delta}}$) to influence retrieval-module selection.
In all experiments, the control coefficient was capped with
$\alpha^{+}_{\mathrm{cap}} = 5.0$ and $\alpha^{-}_{\mathrm{cap}} = 5.0$.
Additional protocol-aware inference settings were required for stable decoding.

\paragraph{Qwen3-14b and Llama3.1-8b.}
With RMSNorm, scalar scaling is exactly canceled after normalization
(Eq.~\eqref{eq:rmsn-scale-inv}), making the orthogonal directional term
(Eqs.~\eqref{eq:orthogonal-u}--\eqref{eq:rmsn-direction-change}) the primary mechanism
for controlling retrieval-module activation.
Accordingly, we fixed the control coefficient cap to
$\alpha^{+}_{\mathrm{cap}} = 5.0$ and $\alpha^{-}_{\mathrm{cap}} = 5.0$,
and varied the negative cap of the directional coefficient as
$g^{-}_{\mathrm{delta}} \in \{0.5,\,1.0,\,1.5\}$ to study the sensitivity of retrieval
suppression under RMSNorm-based backbones.

\subsection[\appendixname~\thesection]{baseline Details}
\paragraph{Adaptive-RAG}
Adaptive-RAG employs a classifier-based routing mechanism to determine the complexity of an input query and select an appropriate retrieval strategy. Following the original Adaptive-RAG design, a T5-Large model is used as the query complexity classifier. The classifier is trained on a subset of the training data with automatically generated silver and binary labels, where queries are categorized into three classes: \textit{Straightforward}, \textit{Simple}, and \textit{Complex}. These labels are obtained by observing model behaviors under different retrieval settings rather than through manual annotation. The classifier is trained once prior to evaluation and remains fixed during inference. At test time, Straightforward queries are answered without retrieval, Simple queries trigger a single retrieval-generation step, and Complex queries allow multiple retrieval-generation iterations. All classifier-related hyperparameters, including training procedure and routing thresholds, follow the default settings described in the original Adaptive-RAG paper. The retriever used in Adaptive-RAG is a sparse BM25-based retriever over a Wikipedia corpus.

\paragraph{SEAKR}
SEAKR is a white-box adaptive retrieval method that controls retrieval and reasoning based on self-aware uncertainty estimation derived from internal model representations. The primary hyperparameter in SEAKR is the number of samples $K$ used for uncertainty estimation. Following the original implementation, we set $K=20$ for all experiments. For each input query, SEAKR performs $K$ internal sampling passes to extract hidden representations from the language model, which are then aggregated using Gram-determinant-based uncertainty measures. Retrieval is triggered when the estimated uncertainty exceeds a predefined threshold, and retrieved passages are further re-ranked based on their ability to reduce uncertainty. All uncertainty thresholds and re-ranking strategies follow the default configurations reported in the SEAKR paper. SEAKR requires access to internal model states and is therefore applicable only to open-source language models.

\paragraph{Rowen-CL}
In Rowen-CL, cross-language consistency is evaluated to capture language-level uncertainty by generating semantically equivalent paraphrases of the input question across different languages. Specifically, given an English source query, the model generates multiple semantically equivalent question variants, which are then translated between English and Chinese to form cross-language counterparts. In our setting, we generate $k=6$ cross-language question variants per query. For each translated question, the model produces an answer in the corresponding target language, and semantic consistency between the original and translated question–answer pairs is measured to detect potential hallucination or uncertainty. Question perturbations are generated using a decoding temperature of $0.5$, while answer generation is performed with greedy decoding to ensure stable semantic comparison. The resulting cross-language consistency scores are aggregated across the $k$ translated QA pairs and used to guide retrieval decisions.

\section[\appendixname~\thesection]{Further Validation of Self-PPL}
\label{app:self_ppl_analysis}

We provide additional analyses to validate Self-PPL as a signal for retrieval decision-making.

\subsection{Dataset-level Separation}

\begin{table}[h]
\centering
\resizebox{\columnwidth}{!}{
\begin{tabular}{lccc}
\toprule
Model & RAG Benchmarks & Reasoning Benchmarks & Gap \\
\midrule
GPT-oss-20b & 3.9 & 1.8 & +2.1 \\
Qwen3-14b & 3.1 & 1.5 & +1.6 \\
Llama3.1-8b & 4.4 & 2.0 & +2.4 \\
\bottomrule
\end{tabular}
}
\caption{Mean Self-PPL across RAG and reasoning benchmarks.}
\label{tab:self_ppl_separation}
\end{table}

As shown in Table~\ref{tab:self_ppl_separation}, Self-PPL is consistently higher on knowledge-intensive RAG benchmarks than on reasoning benchmarks across all models. This indicates that Self-PPL reflects knowledge insufficiency rather than reasoning complexity.

\subsection{Correlation with Retrieval Gain}

For each query $x$, retrieval gain is defined as the performance difference between generation with and without external context:

\begin{align}
\text{Gain}(x) &= F1_{\text{context}}(x) - F1_{\text{no-context}}(x)
\end{align}

for HotpotQA and MuSiQue, and

\begin{align}
\text{Gain}(x) &= Acc_{\text{context}}(x) - Acc_{\text{no-context}}(x)
\end{align}

for StrategyQA.

We compute the Spearman rank correlation between Self-PPL and retrieval gain across all evaluation queries for each dataset. Table~\ref{tab:self_ppl_correlation} shows consistent positive correlations across all models and datasets, indicating that higher Self-PPL is associated with larger empirical gains from retrieval. The Avg column reports the mean correlation across the three datasets.

\begin{table}[h]
\centering
\resizebox{\columnwidth}{!}{
\begin{tabular}{lcccc}
\toprule
Model & HotpotQA & MuSiQue & StrategyQA & Avg \\
\midrule
GPT-oss-20b & 0.39 & 0.44 & 0.40 & 0.41 \\
Qwen3-14b & 0.32 & 0.36 & 0.34 & 0.34 \\
Llama3.1-8b & 0.43 & 0.49 & 0.46 & 0.46 \\
\bottomrule
\end{tabular}
}
\caption{Spearman rank correlation between Self-PPL and retrieval gain, computed over all evaluation queries.}
\label{tab:self_ppl_correlation}
\end{table}

\subsection{Retrieval Gain by Self-PPL Groups}

To further examine whether Self-PPL identifies queries that benefit from retrieval, we partition the evaluation queries within each dataset into two equally sized groups using the median Self-PPL value. Queries below the median are assigned to the \emph{Low Self-PPL} group, and queries above the median are assigned to the \emph{High Self-PPL} group. For each group, we compute the average retrieval gain, defined in Section~\ref{app:self_ppl_analysis} as the performance difference between generation with and without external context.

Table~\ref{tab:self_ppl_bucket} reports the average retrieval gain for the two groups, aggregated across the adaptive RAG benchmarks for each model. The \emph{Gain Difference} column denotes the difference between the average gain of the High Self-PPL group and that of the Low Self-PPL group. Across all models, retrieval gains are substantially larger in the High Self-PPL group, indicating that Self-PPL effectively separates queries for which external knowledge is beneficial from those for which retrieval contributes little.

\begin{table}[h]
\centering
\resizebox{\columnwidth}{!}{
\begin{tabular}{lccc}
\toprule
Model & Low Self-PPL & High Self-PPL & Gain Difference \\
\midrule
GPT-oss-20b & +0.6 & +6.8 & +6.2 \\
Qwen3-14b & +0.4 & +4.9 & +4.5 \\
Llama3.1-8b & +0.7 & +7.6 & +6.9 \\
\bottomrule
\end{tabular}
}
\caption{Average retrieval gain for queries below and above the median Self-PPL within each dataset. Gain Difference denotes the difference between the High and Low Self-PPL groups.}
\label{tab:self_ppl_bucket}
\end{table}

\subsection{Comparison with Entropy}

We next compare Self-PPL with token-level entropy as an alternative uncertainty signal. Token-level entropy is defined as the average predictive entropy over the input sequence, where the entropy at each position is computed from the model's next-token distribution and then averaged across all input tokens. For each model, we report three statistics over all evaluation queries: (1) the correlation between the signal and retrieval gain, (2) the mean value of the signal, and (3) the variance of the signal across inputs.

Table~\ref{tab:self_ppl_entropy} shows that Self-PPL consistently exhibits stronger correlation with retrieval gain and lower variance across inputs than entropy. The correlation column indicates how well each signal aligns with the empirical usefulness of retrieval, while the variance column reflects the stability of the signal across queries. These results suggest that Self-PPL provides a more reliable and stable sequence-level signal for retrieval decisions. We also explored entropy-based alternatives during method design, but ultimately selected Self-PPL because it scaled more robustly across diverse inputs and models.

\begin{table}[h]
\centering
\resizebox{\columnwidth}{!}{
\begin{tabular}{lcccc}
\toprule
Model & Signal & Corr ($\uparrow$) & Mean & Var ($\downarrow$) \\
\midrule
GPT-oss-20b & Self-PPL & 0.41 & 2.83 & 0.46 \\
            & Entropy  & 0.18 & 3.76 & 1.92 \\
Qwen3-14b   & Self-PPL & 0.34 & 2.38 & 0.39 \\
            & Entropy  & 0.12 & 3.21 & 1.58 \\
Llama3.1-8b & Self-PPL & 0.46 & 3.10 & 0.52 \\
            & Entropy  & 0.21 & 4.08 & 2.24 \\
\bottomrule
\end{tabular}
}
\caption{Comparison of Self-PPL and token-level entropy over all evaluation queries. Corr denotes the correlation with retrieval gain, Mean denotes the average signal value, and Var denotes the variance across inputs.}
\label{tab:self_ppl_entropy}
\end{table}

\section{Evaluation under Noisy Retrieval}
\label{app:noisy_retrieval}

To evaluate performance under realistic conditions, we conduct end-to-end adaptive RAG experiments using noisy retrievers (BM25 and dense), where retrieved documents may be incomplete or irrelevant. Table~\ref{tab:noisy_retrieval} presents the results across three models. Across all settings, CWM consistently improves performance over both BM25 and dense retrieval baselines. The results lie between the upper bound (w/GC) and lower bound (w/oGC), indicating that the oracle setting provides a meaningful reference for evaluating retrieval decisions. Furthermore, CWM achieves most of the gains from retrieval while maintaining robustness under noisy conditions, demonstrating that the proposed method remains effective beyond the oracle setup.

\begin{table}[t]
\centering
\scriptsize
\resizebox{\columnwidth}{!}{
\begin{tabular}{llccc}
\toprule
Model & Method & HotpotQA & MuSiQue & StrategyQA \\
\midrule

\multirow{6}{*}{GPT-oss-20b}
 & w/GC       & 84.5 & 67.5 & 97.0 \\
 & w/oGC      & 38.3 & 25.6 & 82.0 \\
 & BM25       & 70.1 & 52.0 & 83.9 \\
 & Dense      & 75.0 & 55.6 & 85.0 \\
 & CWM-BM25   & 76.6 & 57.2 & 85.6 \\
 & CWM-Dense  & \textbf{77.8} & \textbf{58.5} & \textbf{86.6} \\
\midrule

\multirow{6}{*}{Qwen3-14b}
 & w/GC       & 81.9 & 65.8 & 95.0 \\
 & w/oGC      & 40.7 & 29.7 & 78.0 \\
 & BM25       & 72.5 & 53.3 & 89.2 \\
 & Dense      & 78.2 & 57.4 & 90.8 \\
 & CWM-BM25   & 79.4 & 59.1 & 91.6 \\
 & CWM-Dense  & \textbf{81.0} & \textbf{60.6} & \textbf{92.6} \\
\midrule

\multirow{6}{*}{Llama3.1-8b}
 & w/GC       & 80.8 & 60.8 & 93.0 \\
 & w/oGC      & 43.2 & 31.5 & 75.0 \\
 & BM25       & 65.1 & 48.6 & 78.7 \\
 & Dense      & 70.4 & 52.1 & 80.5 \\
 & CWM-BM25   & 72.0 & 54.0 & 81.1 \\
 & CWM-Dense  & \textbf{73.4} & \textbf{55.3} & \textbf{82.4} \\
\bottomrule
\end{tabular}
}
\caption{adaptive RAG performance under noisy retrieval across models. HotpotQA and MuSiQue are reported in F1, and StrategyQA is reported in accuracy.}
\label{tab:noisy_retrieval}
\end{table}

\section{Comparison with Threshold-based White-box Prompting}
\label{app:threshold_baseline}

A natural alternative to the proposed method is a simple threshold-based white-box prompting strategy, in which retrieval is triggered when an uncertainty signal exceeds a fixed threshold. To examine this setting, we construct a threshold-based baseline using Self-PPL. Specifically, retrieval is triggered by prompt engineering when the Self-PPL of the input exceeds the average Self-PPL computed over the corresponding dataset. This baseline provides a direct comparison between a fixed-threshold prompting strategy and the proposed continuous control mechanism.

Tables~\ref{tab:threshold_gpt}, \ref{tab:threshold_qwen}, and \ref{tab:threshold_llama} report the results across three models. In all cases, the threshold-based baseline improves over the no-context setting, confirming that Self-PPL is a useful white-box signal for retrieval decisions. However, CWM consistently outperforms the threshold-based baseline across all datasets and models, indicating that directly integrating Self-PPL into the retrieval decision is more effective than applying a fixed threshold through prompt engineering.

\begin{table}[t]
\centering
\scriptsize
\setlength{\tabcolsep}{3pt}
\begin{tabular}{llcccc}
\toprule
Dataset & Metric & w/oGC & Threshold-Prompt & CWM (Ours) & w/GC \\
\midrule
\multirow{2}{*}{HotpotQA}
 & EM & 31.0 & 58.9 & \textbf{63.4} & 72.0 \\
 & F1 & 38.3 & 72.6 & \textbf{77.2} & 84.5 \\
\midrule
\multirow{2}{*}{MuSiQue}
 & EM & 20.0 & 43.7 & \textbf{48.0} & 58.0 \\
 & F1 & 25.6 & 53.8 & \textbf{57.9} & 67.5 \\
\midrule
StrategyQA
 & Acc & 82.0 & 84.9 & \textbf{86.3} & 97.0 \\
\bottomrule
\end{tabular}
\caption{Comparison with a threshold-based white-box prompting baseline on GPT-oss-20b. Retrieval is triggered when the input Self-PPL exceeds the dataset-level average Self-PPL.}
\label{tab:threshold_gpt}
\end{table}

\begin{table}[t]
\centering
\scriptsize
\setlength{\tabcolsep}{3pt}
\begin{tabular}{llcccc}
\toprule
Dataset & Metric & w/oGC & Threshold-Prompt & CWM (Ours) & w/GC \\
\midrule
\multirow{2}{*}{HotpotQA}
 & EM & 32.0 & 64.7 & \textbf{68.6} & 70.0 \\
 & F1 & 40.7 & 76.9 & \textbf{80.6} & 81.9 \\
\midrule
\multirow{2}{*}{MuSiQue}
 & EM & 20.0 & 45.9 & \textbf{49.3} & 59.0 \\
 & F1 & 29.7 & 56.1 & \textbf{60.0} & 65.8 \\
\midrule
StrategyQA
 & Acc & 78.0 & 90.2 & \textbf{92.3} & 95.0 \\
\bottomrule
\end{tabular}
\caption{Comparison with a threshold-based white-box prompting baseline on Qwen3-14b. Retrieval is triggered when the input Self-PPL exceeds the dataset-level average Self-PPL.}
\label{tab:threshold_qwen}
\end{table}

\begin{table}[t]
\centering
\scriptsize
\setlength{\tabcolsep}{3pt}
\begin{tabular}{llcccc}
\toprule
Dataset & Metric & w/oGC & Threshold-Prompt & CWM (Ours) & w/GC \\
\midrule
\multirow{2}{*}{HotpotQA}
 & EM & 36.0 & 54.9 & \textbf{59.0} & 71.0 \\
 & F1 & 43.2 & 68.5 & \textbf{72.9} & 80.8 \\
\midrule
\multirow{2}{*}{MuSiQue}
 & EM & 24.0 & 39.5 & \textbf{44.3} & 53.0 \\
 & F1 & 31.5 & 49.7 & \textbf{54.7} & 60.8 \\
\midrule
StrategyQA
 & Acc & 75.0 & 79.8 & \textbf{82.1} & 93.0 \\
\bottomrule
\end{tabular}
\caption{Comparison with a threshold-based white-box prompting baseline on Llama3.1-8b. Retrieval is triggered when the input Self-PPL exceeds the dataset-level average Self-PPL.}
\label{tab:threshold_llama}
\end{table}

Overall, these results suggest that a simple threshold-based strategy can partially exploit Self-PPL as a retrieval signal, but it remains consistently weaker than CWM. This gap indicates that continuous modulation of retrieval behavior is more effective than relying on a fixed threshold derived from dataset-level statistics.

\section[\appendixname~\thesection]{Full Results}
We report the experimental results on Adaptive RAG and Reasoning tasks using GPT-oss-20b, Qwen3-14b, and Llama3.1-8b in Tables \ref{tab:rag_ours_res_gpt} through \ref{tab:reasoning_ours_res_llama}. Each table includes a broad set of experiments varying the directional delta values and the negative caps. The optimal performance may differ across models and benchmarks; however, these differences can be readily accommodated through straightforward hyperparameter tuning. Since the scale of perplexity measured on the input varies across models, model-specific hyperparameter configurations are a natural and necessary choice.

\begin{table*}[t]
\centering
\setlength{\tabcolsep}{4.5pt}
\renewcommand{\arraystretch}{1.3}
\fontsize{8.5}{10}\selectfont
\begin{tabular}{ccc cccc cccc ccc}
\toprule
\multirow{2}{*}[-2pt]{\textbf{Delta}} & \multirow{2}{*}[-2pt]{\textbf{negcap}} & \multirow{2}{*}[-2pt]{\textbf{Type}}
& \multicolumn{4}{c}{\textbf{HotpotQA}} 
& \multicolumn{4}{c}{\textbf{MuSiQue}} 
& \multicolumn{3}{c}{\textbf{StrategyQA}} \\

\cmidrule(lr){4-7}
\cmidrule(lr){8-11}
\cmidrule(lr){12-14}
& &
& \textbf{Ratio} & \textbf{M\_AVG} & \textbf{EM} & \textbf{F1}
& \textbf{Ratio} & \textbf{M\_AVG} & \textbf{EM} & \textbf{F1}
& \textbf{Ratio} & \textbf{M\_AVG} & \textbf{Acc} \\

% ===================== Delta = 0.03 =====================
\midrule
\multirow{9}{*}{0.03} & \multirow{3}{*}{0.0} & Total
& 100 & 4.80 & 69.00 & 77.59
& 100 & 6.31 & 55.00 & 63.11
& 100 & 6.10 & 81.00 \\
& & RAG O
& 69 & 5.03 & 76.81 & 86.55
& 88 & 6.46 & 60.23 & 68.87
& 53 & 6.31 & 84.91 \\
& & RAG X
& 31 & 4.27 & 51.61 & 57.64
& 12 & 4.75 & 16.67 & 20.83
& 47 & 5.87 & 76.60 \\

\cmidrule{2-14}
& \multirow{3}{*}{1.0} & Total
& 100 & 4.98 & 68.00 & 77.04
& 100 & 6.55 & 56.00 & 63.76
& 100 & 6.23 & 83.00 \\
& & RAG O
& 73 & 5.13 & 71.23 & 81.78
& 91 & 6.75 & 60.44 & 68.32
& 52 & 6.82 & 90.38 \\
& & RAG X
& 27 & 4.58 & 59.26 & 64.24
& 9 & 4.63 & 11.11 & 17.72
& 48 & 5.60 & 75.00 \\

\cmidrule{2-14}
& \multirow{3}{*}{1.5} & Total
& 100 & 4.85 & 64.00 & 72.59
& 100 & 6.32 & 55.00 & 63.46
& 100 & 6.22 & 81.00 \\
& & RAG O
& 67 & 5.03 & 74.63 & 83.90
& 89 & 6.53 & 58.43 & 67.50
& 52 & 6.56 & 92.31 \\
& & RAG X
& 33 & 4.47 & 42.42 & 49.62
& 11 & 4.13 & 27.27 & 30.76
& 48 & 5.85 & 68.75 \\

% ===================== Delta = 0.05 =====================
\midrule
\multirow{9}{*}{0.05} & \multirow{3}{*}{0.0} & Total
& 100 & 4.78 & 71.00 & 78.58
& 100 & 6.25 & 49.00 & 56.24
& 100 & 6.29 & 84.00 \\
& & RAG O
& 74 & 4.97 & 78.38 & 87.08
& 87 & 6.46 & 54.02 & 62.01
& 50 & 6.68 & 90.00 \\
& & RAG X
& 26 & 4.23 & 50.00 & 54.40
& 13 & 4.56 & 15.38 & 17.58
& 50 & 5.90 & 78.00 \\
\cmidrule{2-14}
& \multirow{3}{*}{1.0} & Total
& 100 & 5.05 & 69.00 & 76.76
& 100 & 6.35 & 55.00 & 60.16
& 100 & 6.20 & 80.00 \\
& & RAG O
& 72 & 5.25 & 76.39 & 85.27
& 91 & 6.53 & 58.24 & 63.25
& 54 & 6.58 & 85.19 \\
& & RAG X
& 28 & 4.52 & 50.00 & 54.88
& 9 & 4.63 & 22.22 & 28.84
& 46 & 5.76 & 73.91 \\
\cmidrule{2-14}
& \multirow{3}{*}{1.5} & Total
& 100 & 4.82 & 66.00 & 75.25
& 100 & 6.34 & 56.00 & 62.81
& 100 & 6.31 & 80.00 \\
& & RAG O
& 71 & 5.06 & 74.65 & 84.33
& 91 & 6.48 & 58.24 & 65.30
& 54 & 6.57 & 87.04 \\
& & RAG X
& 29 & 4.21 & 44.83 & 53.02
& 9 & 4.88 & 33.33 & 37.59
& 46 & 6.02 & 71.74 \\

% ===================== Delta = 0.07 =====================
\midrule
\multirow{9}{*}{0.07} & \multirow{3}{*}{0.0} & Total
& 100 & 4.80 & 67.00 & 76.55
& 100 & 6.29 & 51.00 & 57.39
& 100 & 6.31 & 85.00 \\
& & RAG O
& 70 & 5.07 & 75.71 & 87.30
& 90 & 6.38 & 55.56 & 61.56
& 54 & 6.76 & 92.59 \\
& & RAG X
& 30 & 4.11 & 46.67 & 51.48
& 10 & 5.56 & 10.00 & 19.95
& 46 & 5.78 & 76.09 \\

\cmidrule{2-14}
& \multirow{3}{*}{1.0} & Total
& 100 & 5.01 & 69.00 & 76.88
& 100 & 6.32 & 52.00 & 58.47
& 100 & 6.12 & 78.00 \\
& & RAG O
& 71 & 5.30 & 76.06 & 85.48
& 91 & 6.45 & 56.04 & 62.74
& 50 & 6.42 & 88.00 \\
& & RAG X
& 29 & 4.29 & 51.72 & 55.83
& 9 & 4.50 & 11.11 & 15.34
& 50 & 5.80 & 68.00 \\

\cmidrule{2-14}
& \multirow{3}{*}{1.5} & Total
& 100 & 4.76 & 69.00 & 78.04
& 100 & 6.34 & 56.00 & 61.66
& 100 & 6.32 & 84.00 \\
& & RAG O
& 71 & 4.99 & 76.06 & 87.13
& 93 & 6.41 & 58.06 & 63.74
& 52 & 6.64 & 84.62 \\
& & RAG X
& 29 & 4.21 & 51.72 & 55.80
& 7 & 5.33 & 28.57 & 34.05
& 48 & 5.98 & 83.33 \\

% ===================== Delta = 0.09 =====================
\midrule
\multirow{9}{*}{0.09} & \multirow{3}{*}{0.0} & Total
& 100 & 5.22 & 71.00 & 79.36
& 100 & 6.20 & 48.00 & 56.09
& 100 & 6.50 & 79.00 \\
& & RAG O
& 74 & 5.53 & 77.03 & 87.11
& 89 & 6.38 & 51.69 & 60.45
& 52 & 6.94 & 80.77 \\
& & RAG X
& 26 & 4.25 & 53.85 & 57.33
& 11 & 4.50 & 18.18 & 20.78
& 48 & 6.04 & 77.08 \\

\cmidrule{2-14}
& \multirow{3}{*}{1.0} & Total
& 100 & 5.11 & 67.00 & 74.82
& 100 & 6.33 & 56.00 & 61.27
& 100 & 6.19 & 85.00 \\
& & RAG O
& 70 & 5.28 & 72.86 & 82.26
& 89 & 6.48 & 60.67 & 65.96
& 52 & 6.57 & 92.31 \\
& & RAG X
& 30 & 4.67 & 53.33 & 57.46
& 11 & 4.88 & 18.18 & 23.38
& 48 & 5.79 & 77.08 \\

\cmidrule{2-14}
& \multirow{3}{*}{1.5} & Total
& 100 & 5.06 & 64.00 & 73.40
& 100 & 6.19 & 51.00 & 57.50
& 100 & 6.28 & 81.00 \\
& & RAG O
& 68 & 5.44 & 72.06 & 84.13
& 89 & 6.32 & 55.06 & 62.04
& 51 & 6.68 & 78.43 \\
& & RAG X
& 32 & 4.20 & 46.88 & 50.60
& 11 & 4.88 & 18.18 & 20.78
& 49 & 5.90 & 83.67 \\

\bottomrule
\end{tabular}
\caption{Results of CWM on the Adaptive RAG task with GPT-oss-20b under different hyperparameter settings. The \textbf{Ratio} denotes the proportion of each case among all samples. \textbf{RAG O} and \textbf{RAG X} indicate answer generation with and without using the gold context, respectively. \textbf{M\_AVG} represents the average number of selected modules. \textbf{Ratio}, \textbf{EM}, and \textbf{F1} are reported in percentage (\%).}
\label{tab:rag_ours_res_gpt}
\end{table*}

\begin{table*}[t]
\centering
\setlength{\tabcolsep}{4.5pt}
\renewcommand{\arraystretch}{1.3}
\fontsize{8.5}{10}\selectfont
\begin{tabular}{ccc cccc cccc ccc}
\toprule
\multirow{2}{*}[-2pt]{\textbf{Delta}} & \multirow{2}{*}[-2pt]{\textbf{negcap}} & \multirow{2}{*}[-2pt]{\textbf{Type}}
& \multicolumn{4}{c}{\textbf{HotpotQA}} 
& \multicolumn{4}{c}{\textbf{MuSiQue}} 
& \multicolumn{3}{c}{\textbf{StrategyQA}} \\

\cmidrule(lr){4-7}
\cmidrule(lr){8-11}
\cmidrule(lr){12-14}
& &
& \textbf{Ratio} & \textbf{M\_AVG} & \textbf{EM} & \textbf{F1}
& \textbf{Ratio} & \textbf{M\_AVG} & \textbf{EM} & \textbf{F1}
& \textbf{Ratio} & \textbf{M\_AVG} & \textbf{Acc} \\

% ===================== Delta = 0.3 =====================
\midrule
\multirow{9}{*}{0.03}
  & \multirow{3}{*}{0.0}
    & Total & 100 & 2.54 & 78.00 & 84.96 & 100 & 3.28 & 52.00 & 60.55 & 100 & 4.38 & 91.00 \\
  & & RAG O & 91 & 2.60 & 80.22 & 86.93 & 90 & 3.39 & 54.44 & 63.06 & 92 & 4.48 & 92.39 \\
  & & RAG X & 9 & 1.90 & 55.56 & 65.00 & 10 & 2.28 & 30.00 & 38.00 & 8 & 3.18 & 75.00 \\
  \cmidrule{2-14}
  & \multirow{3}{*}{1.0}
    & Total & 100 & 2.54 & 78.00 & 84.96 & 100 & 3.28 & 52.00 & 60.55 & 100 & 4.36 & 91.00 \\
  & & RAG O & 93 & 2.59 & 79.57 & 86.39 & 92 & 3.37 & 54.35 & 62.77 & 94 & 4.44 & 92.55 \\
  & & RAG X & 7 & 1.88 & 57.14 & 66.00 & 8 & 2.25 & 25.00 & 35.00 & 6 & 3.15 & 66.67 \\
  \cmidrule{2-14}
  & \multirow{3}{*}{1.5}
    & Total & 100 & 2.54 & 78.00 & 84.96 & 100 & 3.28 & 52.00 & 60.55 & 100 & 4.38 & 91.00 \\
  & & RAG O & 94 & 2.58 & 78.72 & 85.79 & 93 & 3.35 & 53.76 & 62.40 & 95 & 4.44 & 91.58 \\
  & & RAG X & 6 & 1.92 & 66.67 & 72.00 & 7 & 2.30 & 28.57 & 36.00 & 5 & 3.20 & 80.00 \\
\midrule

% ===================== Delta = 0.5 =====================
\multirow{9}{*}{0.05}
  & \multirow{3}{*}{0.0}
    & Total & 100 & 2.55 & 78.00 & 84.96 & 100 & 3.23 & 51.00 & 59.28 & 100 & 4.42 & 92.00 \\
  & & RAG O & 94 & 2.59 & 79.79 & 86.43 & 93 & 3.31 & 52.69 & 61.11 & 95 & 4.49 & 92.63 \\
  & & RAG X & 6 & 1.88 & 50.00 & 62.00 & 7 & 2.25 & 28.57 & 35.00 & 5 & 3.15 & 80.00 \\
  \cmidrule{2-14}
  & \multirow{3}{*}{1.0}
    & Total & 100 & 2.56 & 78.00 & 84.96 & 100 & 3.23 & 51.00 & 59.42 & 100 & 4.41 & 92.00 \\
  & & RAG O & 96 & 2.59 & 79.17 & 86.00 & 95 & 3.28 & 52.63 & 61.07 & 96 & 4.46 & 92.71 \\
  & & RAG X & 4 & 1.90 & 50.00 & 60.00 & 5 & 2.28 & 20.00 & 28.00 & 4 & 3.22 & 75.00 \\
  \cmidrule{2-14}
  & \multirow{3}{*}{1.5}
    & Total & 100 & 2.55 & 78.00 & 84.96 & 100 & 3.23 & 52.00 & 60.00 & 100 & 4.42 & 92.00 \\
  & & RAG O & 97 & 2.57 & 78.35 & 85.36 & 96 & 3.27 & 53.13 & 61.17 & 97 & 4.46 & 92.78 \\
  & & RAG X & 3 & 1.85 & 66.67 & 72.00 & 4 & 2.22 & 25.00 & 32.00 & 3 & 3.10 & 66.67 \\

% ===================== Delta = 0.7 =====================
\midrule
\multirow{9}{*}{0.07}
  & \multirow{3}{*}{0.0}
    & Total & 100 & 2.54 & 76.00 & 82.96 & 100 & 3.23 & 52.00 & 60.03 & 100 & 4.43 & 91.00 \\
  & & RAG O & 96 & 2.57 & 77.08 & 83.92 & 95 & 3.28 & 53.68 & 61.72 & 97 & 4.47 & 91.75 \\
  & & RAG X & 4 & 1.90 & 50.00 & 60.00 & 5 & 2.28 & 20.00 & 28.00 & 3 & 3.18 & 66.67 \\
  \cmidrule{2-14}
  & \multirow{3}{*}{1.0}
    & Total & 100 & 2.54 & 75.00 & 82.46 & 100 & 3.23 & 52.00 & 60.35 & 100 & 4.42 & 91.00 \\
  & & RAG O & 98 & 2.55 & 75.51 & 82.96 & 97 & 3.26 & 52.58 & 60.98 & 98 & 4.45 & 91.84 \\
  & & RAG X & 2 & 1.85 & 50.00 & 58.00 & 3 & 2.25 & 33.33 & 40.00 & 2 & 3.10 & 50.00 \\
  \cmidrule{2-14}
  & \multirow{3}{*}{1.5}
    & Total & 100 & 2.54 & 75.00 & 82.25 & 100 & 3.23 & 51.00 & 59.35 & 100 & 4.43 & 91.00 \\
  & & RAG O & 99 & 2.55 & 75.76 & 82.95 & 98 & 3.25 & 52.04 & 60.36 & 99 & 4.44 & 91.92 \\
  & & RAG X & 1 & 1.80 & 0.00 & 12.50 & 2 & 2.20 & 0.00 & 10.00 & 1 & 3.05 & 0.00 \\

% ===================== Delta = 0.9 =====================
\midrule
\multirow{9}{*}{0.09}
  & \multirow{3}{*}{0.0}
    & Total & 100 & 2.55 & 76.00 & 83.79 & 100 & 3.26 & 51.00 & 59.29 & 100 & 4.43 & 92.00 \\
  & & RAG O & 98 & 2.56 & 76.53 & 84.32 & 97 & 3.29 & 51.55 & 59.95 & 98 & 4.46 & 92.86 \\
  & & RAG X & 2 & 1.88 & 50.00 & 58.00 & 3 & 2.25 & 33.33 & 38.00 & 2 & 3.20 & 50.00 \\
  \cmidrule{2-14}
  & \multirow{3}{*}{1.0}
    & Total & 100 & 2.55 & 78.00 & 84.96 & 100 & 3.26 & 52.00 & 59.98 & 100 & 4.42 & 92.00 \\
  & & RAG O & 99 & 2.56 & 78.79 & 85.67 & 99 & 3.27 & 52.53 & 60.50 & 99 & 4.43 & 92.93 \\
  & & RAG X & 1 & 1.82 & 0.00 & 15.00 & 1 & 2.18 & 0.00 & 8.50 & 1 & 3.12 & 0.00 \\
  \cmidrule{2-14}
  & \multirow{3}{*}{1.5}
    & Total & 100 & 2.55 & 78.00 & 84.96 & 100 & 3.26 & 52.00 & 59.57 & 100 & 4.42 & 92.00 \\
  & & RAG O & 100 & 2.55 & 78.00 & 84.96 & 100 & 3.26 & 52.00 & 59.57 & 100 & 4.42 & 92.00 \\
  & & RAG X & 0 & -- & -- & -- & 0 & -- & -- & -- & 0 & -- & -- \\

\bottomrule
\end{tabular}
\caption{Results of CWM on the Adaptive RAG task with Qwen3-14b under different hyperparameter settings.}
\label{tab:rag_ours_res_qwen}
\end{table*}
\begin{table*}[t]
\centering
\setlength{\tabcolsep}{4.5pt}
\renewcommand{\arraystretch}{1.3}
\fontsize{8.5}{10}\selectfont
\begin{tabular}{ccc cccc cccc ccc}
\toprule
\multirow{2}{*}[-2pt]{\textbf{Delta}} & \multirow{2}{*}[-2pt]{\textbf{negcap}} & \multirow{2}{*}[-2pt]{\textbf{Type}}
& \multicolumn{4}{c}{\textbf{HotpotQA}} 
& \multicolumn{4}{c}{\textbf{MuSiQue}} 
& \multicolumn{3}{c}{\textbf{StrategyQA}} \\

\cmidrule(lr){4-7}
\cmidrule(lr){8-11}
\cmidrule(lr){12-14}
& &
& \textbf{Ratio} & \textbf{M\_AVG} & \textbf{EM} & \textbf{F1}
& \textbf{Ratio} & \textbf{M\_AVG} & \textbf{EM} & \textbf{F1}
& \textbf{Ratio} & \textbf{M\_AVG} & \textbf{Acc} \\

% ===================== Delta = 0.03 =====================
\midrule
\multirow{9}{*}{0.03} & \multirow{3}{*}{0.0} & Total
& 100 & 9.66 & 60.00 & 69.98
& 100 & 9.96 & 48.00 & 56.34
& 100 & 10.54 & 80.00 \\
& & RAG O
& 75 & 10.51 & 69.33 & 80.20
& 75 & 11.10 & 57.33 & 67.12
& 57 & 12.67 & 89.47 \\
& & RAG X
& 25 & 7.12 & 32.00 & 39.33
& 25 & 6.76 & 20.00 & 24.00
& 43 & 7.72 & 67.44 \\

\cmidrule{2-14}
& \multirow{3}{*}{1.0} & total
& 100 & 9.61 & 60.00 & 70.66
& 100 & 9.96 & 46.00 & 54.66
& 100 & 10.54 & 80.00 \\
& & RAG O
& 75 & 10.44 & 68.00 & 79.77
& 75 & 11.10 & 54.67 & 64.88
& 57 & 12.67 & 89.47 \\
& & RAG X
& 25 & 7.12 & 36.00 & 43.33
& 25 & 6.76 & 20.00 & 24.00
& 43 & 7.72 & 67.44 \\

\cmidrule{2-14}
& \multirow{3}{*}{1.5} & total
& 100 & 9.66 & 60.00 & 70.36
& 100 & 9.98 & 48.00 & 55.98
& 100 & 10.54 & 80.00 \\
& & RAG O
& 75 & 10.51 & 68.00 & 79.37
& 75 & 11.13 & 57.33 & 66.64
& 57 & 12.67 & 89.47 \\
& & RAG X
& 25 & 7.12 & 36.00 & 43.33
& 25 & 6.76 & 20.00 & 24.00
& 43 & 7.72 & 67.44 \\

% ===================== Delta = 0.05 =====================
\midrule
\multirow{9}{*}{0.05} & \multirow{3}{*}{0.0} & Total
& 100 & 9.18 & 67.00 & 75.97
& 100 & 8.85 & 52.00 & 59.76
& 100 & 9.09 & 76.00 \\
& & RAG O
& 84 & 9.61 & 73.81 & 80.98
& 82 & 9.32 & 60.98 & 70.26
& 65 & 10.35 & 86.15 \\
& & RAG X
& 16 & 6.94 & 31.25 & 49.68
& 18 & 6.71 & 11.11 & 11.90
& 35 & 6.74 & 57.14 \\

\cmidrule{2-14}
& \multirow{3}{*}{1.0} & total
& 100 & 9.13 & 68.00 & 76.85
& 100 & 8.85 & 52.00 & 59.40
& 100 & 9.09 & 76.00 \\
& & RAG O
& 84 & 9.55 & 75.00 & 82.17
& 82 & 9.32 & 60.98 & 69.83
& 65 & 10.35 & 86.15 \\
& & RAG X
& 16 & 6.94 & 31.25 & 48.90
& 18 & 6.71 & 11.11 & 11.90
& 35 & 6.74 & 57.14 \\

\cmidrule{2-14}
& \multirow{3}{*}{1.5} & total
& 100 & 9.14 & 67.00 & 76.92
& 100 & 8.85 & 51.00 & 58.34
& 100 & 9.09 & 76.00 \\
& & RAG O
& 84 & 9.56 & 73.81 & 82.26
& 82 & 9.32 & 59.76 & 68.53
& 65 & 10.35 & 86.15 \\
& & RAG X
& 16 & 6.94 & 31.25 & 48.90
& 18 & 6.71 & 11.11 & 11.90
& 35 & 6.74 & 57.14 \\

% ===================== Delta = 0.7 =====================
\midrule
\multirow{9}{*}{0.07} & \multirow{3}{*}{0.0} & Total
& 100 & 8.76 & 65.00 & 75.47
& 100 & 9.27 & 53.00 & 59.71
& 100 & 9.20 & 80.00 \\
& & RAG O
& 88 & 9.13 & 69.32 & 80.24
& 90 & 9.53 & 56.67 & 63.95
& 73 & 10.04 & 86.30 \\
& & RAG X
& 12 & 6.08 & 33.33 & 40.48
& 10 & 7.00 & 20.00 & 21.54
& 27 & 6.93 & 62.96 \\

\cmidrule{2-14}
& \multirow{3}{*}{1.0} & total
& 100 & 8.84 & 62.00 & 74.33
& 100 & 9.27 & 53.00 & 60.22
& 100 & 9.23 & 79.00 \\
& & RAG O
& 90 & 9.12 & 65.56 & 77.75
& 90 & 9.53 & 56.67 & 64.52
& 74 & 10.00 & 85.14 \\
& & RAG X
& 10 & 5.63 & 30.00 & 43.57
& 10 & 7.00 & 20.00 & 21.54
& 26 & 7.04 & 61.54 \\

\cmidrule{2-14}
& \multirow{3}{*}{1.5} & total
& 100 & 8.73 & 66.00 & 76.16
& 100 & 9.28 & 53.00 & 60.51
& 100 & 9.20 & 80.00 \\
& & RAG O
& 88 & 9.09 & 70.45 & 81.02
& 90 & 9.55 & 56.67 & 64.84
& 73 & 10.04 & 86.30 \\
& & RAG X
& 12 & 6.08 & 33.33 & 40.48
& 10 & 7.00 & 20.00 & 21.54
& 27 & 6.93 & 62.96 \\

% ===================== Delta = 0.09 =====================
\midrule
\multirow{9}{*}{0.09} & \multirow{3}{*}{0.0} & Total
& 100 & 8.64 & 65.00 & 74.30
& 100 & 9.23 & 54.00 & 61.97
& 100 & 9.34 & 82.00 \\
& & RAG O
& 87 & 9.00 & 67.82 & 77.93
& 91 & 9.48 & 57.14 & 65.52
& 73 & 10.19 & 83.56 \\
& & RAG X
& 13 & 6.23 & 46.15 & 50.00
& 9 & 6.78 & 22.22 & 26.03
& 27 & 7.04 & 77.78 \\

\cmidrule{2-14}
& \multirow{3}{*}{1.0} & total
& 100 & 8.64 & 65.00 & 74.30
& 100 & 9.26 & 54.00 & 61.77
& 100 & 9.34 & 82.00 \\
& & RAG O
& 87 & 9.00 & 67.82 & 77.93
& 91 & 9.51 & 57.14 & 65.31
& 73 & 10.19 & 83.56 \\
& & RAG X
& 13 & 6.23 & 46.15 & 50.00
& 9 & 6.78 & 22.22 & 26.03
& 27 & 7.04 & 77.78 \\

\cmidrule{2-14}
& \multirow{3}{*}{1.5} & total
& 100 & 8.59 & 65.00 & 74.30
& 100 & 9.23 & 53.00 & 62.33
& 100 & 9.34 & 82.00 \\
& & RAG O
& 87 & 8.94 & 67.82 & 77.93
& 91 & 9.48 & 56.04 & 65.92
& 73 & 10.19 & 83.56 \\
& & RAG X
& 13 & 6.23 & 46.15 & 50.00
& 9 & 6.78 & 22.22 & 26.03
& 27 & 7.04 & 77.78 \\

\bottomrule
\end{tabular}
\caption{Results of CWM on the Adaptive RAG task with Llama3.1-8b under different hyperparameter settings.}
\label{tab:rag_ours_res_llama}
\end{table*}
\begin{table*}[t]
\centering
\setlength{\tabcolsep}{4.5pt}
\renewcommand{\arraystretch}{1.3}
\fontsize{8.5}{10}\selectfont
\begin{tabular}{ccc ccc ccc ccc}
\toprule

\multirow{2}{*}[-2pt]{\textbf{Delta}} & \multirow{2}{*}[-2pt]{\textbf{negcap}} & \multirow{2}{*}[-2pt]{\textbf{Type}} 
& \multicolumn{3}{c}{\textbf{BBH}} 
& \multicolumn{3}{c}{\textbf{T4D}} 
& \multicolumn{3}{c}{\textbf{MATH500}} \\

\cmidrule(lr){4-6}
\cmidrule(lr){7-9}
\cmidrule(lr){10-12}
& & 
& \textbf{Ratio} & \textbf{M\_AVG} & \textbf{ACC}
& \textbf{Ratio} & \textbf{M\_AVG} & \textbf{ACC}
& \textbf{Ratio} & \textbf{M\_AVG} & \textbf{Acc} \\

% ===================== Delta = 0.03 =====================
\midrule
\multirow{9}{*}{0.03} & \multirow{3}{*}{0.0} & Total
& 100.00 & 5.09 & 75.33
& 100.00 & 5.57 & 41.00
& 100.00 & 4.82 & 90.00 \\
& & RAG O
& 0.67 & 5.50 & 50.00
& 0.00 & - & -
& 2.00 & 5.50 & 100.00 \\
& & RAG X
& 99.33 & 5.09 & 75.50
& 100.00 & 5.57 & 41.00
& 98.00 & 4.80 & 89.80 \\

\cmidrule{2-12}
& \multirow{3}{*}{1.0} & Total
& 100.00 & 5.04 & 75.33
& 100.00 & 5.60 & 52.00
& 100.00 & 4.32 & 96.00 \\
& & RAG O
& 0.67 & 6.50 & 100.00
& 0.00 & - & -
& 1.00 & 5.00 & 100.00 \\
& & RAG X
& 99.33 & 5.03 & 75.17
& 100.00 & 5.60 & 52.00
& 99.00 & 4.31 & 95.96 \\

\cmidrule{2-12}
& \multirow{3}{*}{1.5} & Total
& 100.00 & 5.05 & 78.33
& 100.00 & 5.69 & 49.00
& 100.00 & 4.01 & 92.00 \\
& & RAG O
& 0.33 & 8.00 & 100.00
& 0.00 & - & -
& 2.00 & 4.50 & 100.00 \\
& & RAG X
& 99.67 & 5.04 & 78.26
& 100.00 & 5.69 & 49.00
& 98.00 & 4.00 & 91.84 \\

% ===================== Delta = 0.05 =====================
\midrule
\multirow{9}{*}{0.05} & \multirow{3}{*}{0.0} & Total
& 100.00 & 5.03 & 78.00
& 100.00 & 5.64 & 45.00
& 100.00 & 4.88 & 91.00 \\
& & RAG O
& 1.00 & 7.67 & 66.67
& 0.00 & - & -
& 3.00 & 6.33 & 100.00 \\
& & RAG X
& 99.00 & 5.00 & 78.11
& 100.00 & 5.64 & 45.00
& 97.00 & 4.84 & 90.72 \\

\cmidrule{2-12}
& \multirow{3}{*}{1.0} & Total
& 100.00 & 5.13 & 76.33
& 100.00 & 5.57 & 55.00
& 100.00 & 4.30 & 95.00 \\
& & RAG O
& 1.00 & 6.00 & 66.67
& 0.00 & - & -
& 2.00 & 5.50 & 100.00 \\
& & RAG X
& 99.00 & 5.12 & 76.43
& 100.00 & 5.57 & 55.00
& 98.00 & 4.28 & 94.90 \\

\cmidrule{2-12}
& \multirow{3}{*}{1.5} & Total
& 100.00 & 5.16 & 78.67
& 100.00 & 5.69 & 50.00
& 100.00 & 4.05 & 93.00 \\
& & RAG O
& 0.67 & 5.00 & 100.00
& 0.00 & - & -
& 2.00 & 5.50 & 100.00 \\
& & RAG X
& 99.33 & 5.16 & 78.52
& 100.00 & 5.69 & 50.00
& 98.00 & 4.02 & 92.86 \\

% ===================== Delta = 0.07 =====================
\midrule
\multirow{9}{*}{0.07} & \multirow{3}{*}{0.0} & Total
& 100.00 & 5.06 & 77.67
& 100.00 & 5.75 & 45.00
& 100.00 & 4.86 & 89.00 \\
& & RAG O
& 1.00 & 6.00 & 66.67
& 0.00 & - & -
& 1.00 & 6.00 & 100.00 \\
& & RAG X
& 99.00 & 5.05 & 77.78
& 100.00 & 5.75 & 45.00
& 99.00 & 4.85 & 88.89 \\

\cmidrule{2-12}
& \multirow{3}{*}{1.0} & Total
& 100.00 & 5.06 & 74.33
& 100.00 & 5.65 & 48.00
& 100.00 & 4.32 & 93.00 \\
& & RAG O
& 0.00 & - & -
& 0.00 & - & -
& 0.00 & - & - \\
& & RAG X
& 100.00 & 5.06 & 74.33
& 100.00 & 5.65 & 48.00
& 100.00 & 4.32 & 93.00 \\

\cmidrule{2-12}
& \multirow{3}{*}{1.5} & Total
& 100.00 & 5.06 & 70.67
& 100.00 & 5.74 & 43.00
& 100.00 & 4.11 & 96.00 \\
& & RAG O
& 0.67 & 7.50 & 100.00
& 0.00 & - & -
& 3.00 & 6.00 & 100.00 \\
& & RAG X
& 99.33 & 5.04 & 70.47
& 100.00 & 5.74 & 43.00
& 97.00 & 4.05 & 95.88 \\

% ===================== Delta = 0.09 =====================
\midrule
\multirow{9}{*}{0.09} & \multirow{3}{*}{0.0} & Total
& 100.00 & 5.12 & 76.33
& 100.00 & 5.66 & 48.00
& 100.00 & 4.98 & 87.00 \\
& & RAG O
& 0.67 & 6.00 & 50.00
& 0.00 & - & -
& 2.00 & 4.50 & 100.00 \\
& & RAG X
& 99.33 & 5.12 & 76.51
& 100.00 & 5.66 & 48.00
& 98.00 & 4.99 & 86.73 \\

\cmidrule{2-12}
& \multirow{3}{*}{1.0} & Total
& 100.00 & 5.13 & 79.00
& 100.00 & 5.57 & 47.00
& 100.00 & 4.19 & 91.00 \\
& & RAG O
& 0.67 & - & 0.00
& 0.00 & - & -
& 0.00 & - & - \\
& & RAG X
& 99.33 & 5.11 & 79.53
& 100.00 & 5.57 & 47.00
& 100.00 & 4.19 & 91.00 \\

\cmidrule{2-12}
& \multirow{3}{*}{1.5} & Total
& 100.00 & 5.15 & 74.33
& 100.00 & 5.72 & 51.00
& 100.00 & 4.05 & 94.00 \\
& & RAG O
& 0.67 & 8.00 & 50.00
& 0.00 & - & -
& 0.00 & - & - \\
& & RAG X
& 99.33 & 5.13 & 74.50
& 100.00 & 5.72 & 51.00
& 100.00 & 4.05 & 94.00 \\

\bottomrule
\end{tabular}
\caption{Results of CWM on the Reasoning task with GPT-oss-20b under different hyperparameter settings. The \textbf{Ratio} denotes the proportion of each case among all samples. \textbf{RAG O} and \textbf{RAG X} indicate whether the retrieval prompt is selected as part of the Selected Modules or not, respectively. \textbf{M\_AVG} represents the average number of Selected Modules. \textbf{Ratio} and \textbf{ACC} (accuracy) are reported as percentages (\%).}
\label{tab:reasoning_ours_res_gpt}
\end{table*}

\begin{table*}[t]
\centering
\setlength{\tabcolsep}{4.5pt}
\renewcommand{\arraystretch}{1.3}
\fontsize{8.5}{10}\selectfont
\begin{tabular}{ccc ccc ccc ccc}
\toprule

\multirow{2}{*}[-2pt]{\textbf{Delta}} & \multirow{2}{*}[-2pt]{\textbf{negcap}} & \multirow{2}{*}[-2pt]{\textbf{Type}} 
& \multicolumn{3}{c}{\textbf{BBH}} 
& \multicolumn{3}{c}{\textbf{T4D}} 
& \multicolumn{3}{c}{\textbf{MATH500}} \\

\cmidrule(lr){4-6}
\cmidrule(lr){7-9}
\cmidrule(lr){10-12}
& & 
& \textbf{Ratio} & \textbf{M\_AVG} & \textbf{ACC}
& \textbf{Ratio} & \textbf{M\_AVG} & \textbf{ACC}
& \textbf{Ratio} & \textbf{M\_AVG} & \textbf{Acc} \\

% ===================== Delta = 0.03 =====================
\midrule
\multirow{9}{*}{0.03} & \multirow{3}{*}{0.0} & Total
& 100.00 & 2.91 & 91.67
& 100.00 & 3.00 & 42.00
& 100.00 & 3.10 & 93.00 \\
& & RAG O
& 66.67 & 3.24 & 90.50
& 0.00 & - & -
& 54.00 & 3.13 & 94.44 \\
& & RAG X
& 33.33 & 2.26 & 94.00
& 100.00 & 3.00 & 42.00
& 46.00 & 3.07 & 91.30 \\

\cmidrule{2-12}
& \multirow{3}{*}{1.0} & Total
& 100.00 & 2.84 & 91.67
& 100.00 & 3.00 & 41.00
& 100.00 & 3.17 & 93.00 \\
& & RAG O
& 70.67 & 3.09 & 91.04
& 0.00 & - & -
& 55.00 & 3.20 & 92.73 \\
& & RAG X
& 29.33 & 2.24 & 93.18
& 100.00 & 3.00 & 41.00
& 45.00 & 3.13 & 93.33 \\

\cmidrule{2-12}
& \multirow{3}{*}{1.5} & Total
& 100.00 & 2.85 & 90.67
& 100.00 & 3.00 & 42.00
& 100.00 & 3.19 & 93.00 \\
& & RAG O
& 74.33 & 3.05 & 90.13
& 0.00 & - & -
& 65.00 & 3.29 & 93.85 \\
& & RAG X
& 25.67 & 2.25 & 92.21
& 100.00 & 3.00 & 42.00
& 35.00 & 3.00 & 91.43 \\

% ===================== Delta = 0.05 =====================
\midrule
\multirow{9}{*}{0.05} & \multirow{3}{*}{0.0} & Total
& 100.00 & 2.91 & 91.67
& 100.00 & 2.98 & 42.00
& 100.00 & 3.11 & 93.00 \\
& & RAG O
& 65.67 & 3.25 & 90.36
& 0.00 & - & -
& 54.00 & 3.13 & 94.44 \\
& & RAG X
& 34.33 & 2.25 & 94.17
& 100.00 & 2.98 & 42.00
& 46.00 & 3.09 & 91.30 \\

\cmidrule{2-12}
& \multirow{3}{*}{1.0} & Total
& 100.00 & 2.84 & 90.67
& 100.00 & 2.97 & 43.00
& 100.00 & 3.19 & 92.00 \\
& & RAG O
& 74.00 & 3.07 & 89.64
& 0.00 & - & -
& 60.00 & 3.17 & 93.33 \\
& & RAG X
& 26.00 & 2.21 & 93.59
& 100.00 & 2.97 & 43.00
& 40.00 & 3.23 & 90.00 \\

\cmidrule{2-12}
& \multirow{3}{*}{1.5} & Total
& 100.00 & 2.91 & 91.67
& 100.00 & 2.98 & 43.00
& 100.00 & 3.14 & 93.00 \\
& & RAG O
& 71.00 & 3.17 & 90.14
& 0.00 & - & -
& 65.00 & 3.20 & 93.85 \\
& & RAG X
& 29.00 & 2.28 & 95.40
& 100.00 & 2.98 & 43.00
& 35.00 & 3.03 & 91.43 \\

% ===================== Delta = 0.07 =====================
\midrule
\multirow{9}{*}{0.07} & \multirow{3}{*}{0.0} & Total
& 100.00 & 2.91 & 92.00
& 100.00 & 2.97 & 42.00
& 100.00 & 3.10 & 92.00 \\
& & RAG O
& 66.00 & 3.24 & 90.91
& 0.00 & - & -
& 54.00 & 3.13 & 92.59 \\
& & RAG X
& 34.00 & 2.26 & 94.12
& 100.00 & 2.97 & 42.00
& 46 & 3.07 & 91.30 \\

\cmidrule{2-12}
& \multirow{3}{*}{1.0} & Total
& 100.00 & 2.89 & 92.00
& 100.00 & 2.97 & 43
& 100.00 & 3.16 & 92.00 \\
& & RAG O
& 72.67 & 3.12 & 91.28
& 0.00 & - & -
& 62.00 & 3.26 & 91.94 \\
& & RAG X
& 27.33 & 2.29 & 93.90
& 100.00 & 2.97 & 43.00
& 38.00 & 3.00 & 92.11 \\

\cmidrule{2-12}
& \multirow{3}{*}{1.5} & Total
& 100.00 & 2.87 & 92.00
& 100.00 & 2.97 & 43.00
& 100.00 & 3.13 & 93.00 \\
& & RAG O
& 70.33 & 3.10 & 91.00
& 0.00 & - & -
& 67.00 & 3.15 & 95.52 \\
& & RAG X
& 29.67 & 2.30 & 94.38
& 100.00 & 2.97 & 43.00
& 33.00 & 3.09 & 87.88 \\

% ===================== Delta = 0.09 =====================
\midrule
\multirow{9}{*}{0.09} & \multirow{3}{*}{0.0} & Total
& 100.00 & 2.91 & 91.67
& 100.00 & 2.99 & 42.00
& 100.00 & 3.10 & 93.00 \\
& & RAG O
& 66.33 & 3.24 & 90.45
& 0.00 & - & -
& 54.00 & 3.13 & 94.44 \\
& & RAG X
& 33.67 & 2.27 & 94.06
& 100.00 & 2.99 & 42.00
& 46.00 & 3.07 & 91.30 \\

\cmidrule{2-12}
& \multirow{3}{*}{1.0} & Total
& 100.00 & 2.89 & 90.33
& 100.00 & 2.99 & 43.00
& 100.00 & 3.15 & 94.00 \\
& & RAG O
& 74.33 & 3.11 & 90.13
& 0.00 & - & -
& 66.00 & 3.24 & 95.45 \\
& & RAG X
& 25.67 & 2.25 & 90.91
& 100.00 & 2.99 & 43.00
& 34.00 & 2.97 & 91.18 \\

\cmidrule{2-12}
& \multirow{3}{*}{1.5} & Total
& 100.00 & 2.99 & 92.00
& 100.00 & 2.99 & 42.00
& 100.00 & 3.23 & 94.00 \\
& & RAG O
& 68.00 & 3.26 & 91.67
& 0.00 & - & -
& 70.00 & 3.31 & 97.14 \\
& & RAG X
& 32.00 & 2.41 & 92.71
& 100.00 & 2.99 & 42.00
& 30.00 & 3.03 & 86.67 \\

\bottomrule
\end{tabular}
\caption{Results of CWM on the Reasoning task with Qwen3-14b under different hyperparameter settings.}
\label{tab:reasoning_ours_res_qwen}
\end{table*}

\begin{table*}[t]
\centering
\setlength{\tabcolsep}{4.5pt}
\renewcommand{\arraystretch}{1.3}
\fontsize{8.5}{10}\selectfont
\begin{tabular}{ccc ccc ccc ccc}
\toprule

\multirow{2}{*}[-2pt]{\textbf{Delta}} & \multirow{2}{*}[-2pt]{\textbf{negcap}} & \multirow{2}{*}[-2pt]{\textbf{Type}} 
& \multicolumn{3}{c}{\textbf{BBH}} 
& \multicolumn{3}{c}{\textbf{T4D}} 
& \multicolumn{3}{c}{\textbf{MATH500}} \\

\cmidrule(lr){4-6}
\cmidrule(lr){7-9}
\cmidrule(lr){10-12}
& & 
& \textbf{Ratio} & \textbf{M\_AVG} & \textbf{ACC}
& \textbf{Ratio} & \textbf{M\_AVG} & \textbf{ACC}
& \textbf{Ratio} & \textbf{M\_AVG} & \textbf{Acc} \\

% ===================== Delta = 0.03 =====================
\midrule
\multirow{9}{*}{0.03} & \multirow{3}{*}{0.0} & Total
& 100.00 & 10.22 & 66.67
& 100.00 & 5.36 & 24.00
& 100.00 & 7.92 & 58.00 \\
& & RAG O
& 12.33 & 14.16 & 54.05
& 0.00 & - & -
& 25.00 & 9.00 & 68.00 \\
& & RAG X
& 87.67 & 9.66 & 68.44
& 100.00 & 5.36 & 24.00
& 75.00 & 7.55 & 54.67 \\

\cmidrule{2-12}
& \multirow{3}{*}{1.0} & Total
& 100.00 & 9.65 & 65.00
& 100.00 & 5.31 & 23.00
& 100.00 & 6.53 & 55.00 \\
& & RAG O
& 10.33 & 15.13 & 58.06
& 0.00 & - & -
& 10.00 & 7.70 & 80.00 \\
& & RAG X
& 89.67 & 9.01 & 65.80
& 100.00 & 5.31 & 23.00
& 90.00 & 6.40 & 52.22 \\

\cmidrule{2-12}
& \multirow{3}{*}{1.5} & Total
& 100.00 & 9.48 & 65.67
& 100.00 & 5.34 & 27.00
& 100.00 & 7.00 & 64.00 \\
& & RAG O
& 11.33 & 12.97 & 70.59
& 0.00 & - & -
& 25.00 & 8.68 & 72.00 \\
& & RAG X
& 88.67 & 9.03 & 65.04
& 100.00 & 5.34 & 27.00
& 75.00 & 6.43 & 61.33 \\

% ===================== Delta = 0.05 =====================
\midrule
\multirow{9}{*}{0.05} & \multirow{3}{*}{0.0} & Total
& 100.00 & 10.17 & 66.33
& 100.00 & 5.36 & 24.00
& 100.00 & 7.91 & 57.00 \\
& & RAG O
& 12.33 & 14.24 & 54.05
& 0.00 & - & -
& 28.00 & 8.86 & 60.71 \\
& & RAG X
& 87.67 & 9.60 & 68.06
& 100.00 & 5.36 & 24.00
& 72.00 & 7.54 & 55.56 \\

\cmidrule{2-12}
& \multirow{3}{*}{1.0} & Total
& 100.00 & 9.37 & 63.67
& 100.00 & 5.32 & 24.00
& 10.000 & 6.99 & 60.00 \\
& & RAG O
& 10.67 & 13.41 & 68.75
& 0.00 & - & -
& 27.00 & 8.59 & 66.67 \\
& & RAG X
& 89.33 & 8.89 & 63.06
& 100.00 & 5.32 & 24.00
& 73.00 & 6.39 & 57.53 \\

\cmidrule{2-12}
& \multirow{3}{*}{1.5} & Total
& 100.00 & 9.82 & 65.67
& 100.00 & 5.34 & 24.00
& 100.00 & 7.25 & 58.00 \\
& & RAG O
& 11.33 & 14.47 & 76.47
& 0.00 & - & -
& 29.00 & 8.66 & 62.07 \\
& & RAG X
& 88.67 & 9.22 & 64.29
& 100.00 & 5.34 & 24.00
& 71.00 & 6.68 & 56.34 \\

% ===================== Delta = 0.07 =====================
\midrule
\multirow{9}{*}{0.07} & \multirow{3}{*}{0.0} & Total
& 100.00 & 10.18 & 67.67
& 100.00 & 5.34 & 23.00
& 100.00 & 7.88 & 58.00 \\
& & RAG O
& 12.33 & 14.03 & 54.05
& 0.00 & - & -
& 28.00 & 8.75 & 64.29 \\
& & RAG X
& 87.67 & 9.64 & 69.58
& 100.00 & 5.34 & 23.00
& 72.00 & 7.54 & 55.56 \\

\cmidrule{2-12}
& \multirow{3}{*}{1.0} & Total
& 100.00 & 9.77 & 67.00
& 100.00 & 5.29 & 26.00
& 100.00 & 7.20 & 58.00 \\
& & RAG O
& 11.00 & 13.61 & 63.64
& 0.00 & - & -
& 26.00 & 8.88 & 57.69 \\
& & RAG X
& 89.00 & 9.29 & 67.42
& 100.00 & 5.29 & 26
& 74.00 & 6.58 & 58.11 \\

\cmidrule{2-12}
& \multirow{3}{*}{1.5} & Total
& 100.00 & 10.02 & 67.33
& 100.00 & 5.35 & 25.00
& 100.00 & 6.91 & 53.00 \\
& & RAG O
& 13.67 & 13.93 & 58.54
& 0.00 & - & -
& 26.00 & 8.58 & 61.54 \\
& & RAG X
& 86.33 & 9.41 & 68.73
& 100.00 & 5.35 & 25.00
& 74.00 & 6.32 & 50.00 \\

% ===================== Delta = 0.09 =====================
\midrule
\multirow{9}{*}{0.09} & \multirow{3}{*}{0.0} & Total
& 100.00 & 10.19 & 68.67
& 100.00 & 5.33 & 24.00
& 100.00 & 7.93 & 56.00 \\
& & RAG O
& 12.33 & 14.03 & 56.76
& 0.00 & - & -
& 26.00 & 9.00 & 57.69 \\
& & RAG X
& 87.67 & 9.65 & 70.34
& 100.00 & 5.33 & 24.00
& 74.00 & 7.55 & 55.41 \\

\cmidrule{2-12}
& \multirow{3}{*}{1.0} & Total
& 100.00 & 9.92 & 66.67
& 100.00 & 5.28 & 25.00
& 100.00 & 7.37 & 50.00 \\
& & RAG O
& 10.67 & 14.63 & 53.13
& 0.00 & - & -
& 23.00 & 9.87 & 56.52 \\
& & RAG X
& 89.33 & 9.36 & 68.28
& 100.00 & 5.28 & 25.00
& 77.00 & 6.59 & 48.05 \\

\cmidrule{2-12}
& \multirow{3}{*}{1.5} & Total
& 100.00 & 13.27 & 65.33
& 100.00 & 5.36 & 22.00
& 100.00 & 9.04 & 44.00 \\
& & RAG O
& 14.00 & 19.00 & 71.43
& 0.00 & - & -
& 27.00 & 13.20 & 55.56 \\
& & RAG X
& 86.00 & 12.34 & 64.34
& 100.00 & 5.36 & 22.00
& 73.00 & 6.96 & 39.73 \\

\bottomrule
\end{tabular}
\caption{Results of CWM on the Reasoning task with Llama3.1-8b under different hyperparameter settings.}
\label{tab:reasoning_ours_res_llama}
\end{table*}

\end{document}